\documentclass[11pt]{article}

\usepackage[table]{xcolor}

\definecolor{orange}{HTML}{FF7F0E}
\definecolor{green}{HTML}{2CA02C}
\definecolor{darkred}{HTML}{cc0808}

\newcommand*\rot{\rotatebox{90}}

\usepackage{authblk}
\usepackage{acl}

\usepackage{float}
\usepackage{graphicx}

\usepackage{tcolorbox}
\usepackage{amsmath}
\usepackage{newtxmath}
\usepackage[normalem]{ulem}

\usepackage{times}
\usepackage{latexsym}
\usepackage{enumitem}
\usepackage{multirow}
\usepackage{booktabs}
\usepackage{tabularx}
\usepackage{array}

\usepackage[T1]{fontenc}

\usepackage[utf8]{inputenc}

\usepackage{siunitx}
\usepackage{array}
\usepackage{microtype}
\usepackage{subcaption}

\usepackage{cleveref}
\usepackage{booktabs}

\usepackage[frozencache=true,cachedir=minted-cache]{minted} 
\usepackage{graphicx}
\usepackage[inkscapearea=drawing]{svg}
\usepackage[colorinlistoftodos]{todonotes}
\usepackage{multirow}
\usepackage{pbox}
\usepackage{xspace}
\usepackage{stfloats}

\usepackage{url}
\usepackage{tikz}
\usetikzlibrary{fadings,shapes.arrows,shadows} 

  \tikzfading[
  name=arrowfading,
  top color=transparent!0,
  bottom color=transparent!95
]

\usepackage[breaklinks]{hyperref}

\usepackage{svg}
\svgpath{{./figures/}}
\usepackage{wrapfig}

\title{CATfOOD: Counterfactual Augmented Training for Improving Out-of-Domain Performance and Calibration}

\author{\textbf{Rachneet Sachdeva, Martin Tutek, Iryna Gurevych} \\
    Ubiquitous Knowledge Processing Lab (UKP Lab) \\
    Department of Computer Science and Hessian Center for AI (hessian.AI)\\
    Technical University of Darmstadt \\
    \url{www.ukp.tu-darmstadt.de} \\
}

\begin{document}
\maketitle

\begin{abstract}
In recent years, large language models (LLMs) have shown remarkable capabilities at scale, particularly at generating text conditioned on a prompt. 
In our work, we investigate the use of LLMs to augment training data of smaller language models~(SLMs) with automatically generated counterfactual~(CF) instances -- i.e. minimally altered inputs -- in order to improve out-of-domain~(OOD) performance of SLMs in the extractive question answering~(QA) setup.
We show that, across various LLM generators, such data augmentation consistently enhances OOD performance and improves model calibration for both confidence-based and rationale-augmented calibrator models.
Furthermore, these performance improvements correlate with higher diversity of CF instances in terms of their surface form and semantic content. %
Finally, we show that CF augmented models which are easier to calibrate also exhibit much lower entropy when assigning importance, indicating that rationale-augmented calibrators prefer concise explanations.\footnote{We make our code available at: \href{https://github.com/UKPLab/CATfOOD/}{github.com/CATfOOD}}

\end{abstract}

\section{Introduction}
Ever since their introduction to the field of NLP, large language models~(LLMs) have shown exceptional performance across a wide array of applications~(\citealt{devlin-etal-2019-bert, NEURIPS2020_1457c0d6, DBLP:journals/corr/abs-2206-07682}; \textit{inter alia}).
LLMs have frequently been utilized to enhance reasoning capabilities of smaller models~\cite{DBLP:journals/corr/abs-2210-06726}, generate counterfactuals~(CF) -- minimally perturbed input instances -- for data augmentation~\cite{DBLP:journals/corr/abs-2206-13757, paranjape-etal-2022-retrieval}, and have shown remarkable generalization capabilities, performing well on various tasks such as question answering~(QA), complex reasoning, and code generation~\cite{DBLP:conf/iclr/WeiBZGYLDDL22,  DBLP:journals/corr/abs-2204-06745, DBLP:journals/corr/abs-2302-13971}. 
On the other hand, comparatively small language models (SLMs) such as BERT~\cite{devlin-etal-2019-bert} perform well on task specific data but their performance drops with a change in the data distribution~\cite{DBLP:conf/icml/KohSMXZBHYPGLDS21, he2023preserving} and they are frequently poorly calibrated, exhibiting under- or overconfidence in their predictions \cite{desai-durrett-2020-calibration, kong-etal-2020-calibrated, guo-etal-2021-overview, jiang-etal-2021-know}.
In our paper, we examine how data augmentation with CFs of varying diversity improves out-of-domain~(OOD) performance and model calibration of SLMs. For comparability to previous work, we perform our experiments in the extractive QA domain, but we believe our findings could generalize to other tasks given the remarkable versatility exhibited by LLMs~\cite{DBLP:conf/iclr/WeiBZGYLDDL22}.

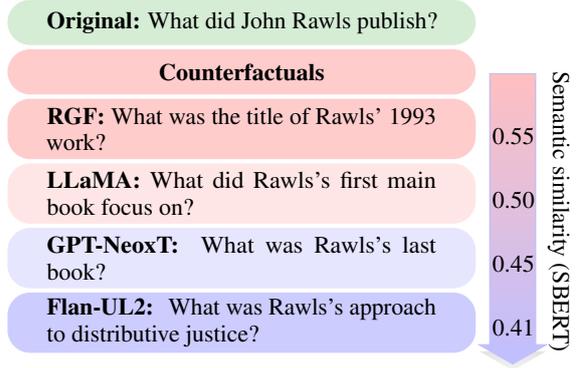
\begin{figure}[!t]
\centering
\begin{tikzpicture}
  \node[] (box) {
    \begin{minipage}{0.8\linewidth}

    \begin{tcolorbox}[colback=green!20, colframe=green!20, arc=8pt, boxrule=0.5pt, width=\linewidth, height=0.6cm, valign=center, fontupper=\small]
        {\textbf{Original:} What did John Rawls publish?}
        \end{tcolorbox}

    \vspace{-12pt}
    \begin{tcolorbox}[colback=red!20, colframe=red!20, arc=8pt, boxrule=0.5pt, width=\linewidth, height=0.6cm, valign=center, fontupper=\small]
        \centering\textbf{Counterfactuals}
        \end{tcolorbox}

    \vspace{-12pt}
    \begin{tcolorbox}[colback=red!20, colframe=red!20, arc=8pt, boxrule=0.5pt, width=\linewidth, height=0.8cm, valign=center, fontupper=\small]
        {\textbf{RGF:} What was the title of Rawls' 1993 work?}
        \end{tcolorbox}

    \vspace{-12pt}
    \begin{tcolorbox}[colback=red!10, colframe=red!10, arc=8pt, boxrule=0.5pt, width=\linewidth, height=0.8cm, valign=center, fontupper=\small]
        {\textbf{LLaMA: }What did Rawls's first main book focus on?}
        \end{tcolorbox}

    \vspace{-12pt}
    \begin{tcolorbox}[colback=blue!10, colframe=blue!10, arc=8pt, boxrule=0.5pt, width=\linewidth, height=0.8cm, valign=center, fontupper=\small]
        {\textbf{GPT-NeoxT: } What was Rawls's last book?}
        \end{tcolorbox}

    \vspace{-12pt}
    \begin{tcolorbox}[colback=blue!20, colframe=blue!20, arc=8pt, boxrule=0.5pt, width=\linewidth, height=0.8cm, valign=center, fontupper=\small]
        {\textbf{Flan-UL2: } What was Rawls's approach to distributive justice?}
        \end{tcolorbox}

    \end{minipage}
  };

\node[
font=\footnotesize,
single arrow,
single arrow head extend=4pt,
single arrow tip angle=120,
shape border rotate=270,
draw=blue!25,
inner sep=2pt,
top color=red!25,
bottom color=blue!25, 
general shadow={
    fill=black,
    shadow yshift=-0.5ex,
    path fading=arrowfading
},
minimum height=3.85cm, 
align=center, 
text width=0.45cm 
] at ([xshift=10pt, yshift=-84pt] box.north east) {};

\node[anchor=north east, font=\small] at ([xshift=22pt, yshift=22pt] box.east) {0.55};
\node[anchor=north east, font=\small] at ([xshift=22pt, yshift=-2pt] box.east) {0.50};
\node[anchor=north east, font=\small] at ([xshift=22pt, yshift=-26pt] box.east) {0.45};
\node[anchor=north east, font=\small] at ([xshift=22pt, yshift=-50pt] box.east) {0.41};

\node[anchor=north east, font=\small, rotate=270] at ([xshift=35pt, yshift=-70pt, ] box.east) {Semantic similarity (SBERT)};

\end{tikzpicture}
\caption{An illustration of the counterfactual samples~(purple) for the input question~(green) produced by the RGF baseline and our approaches using LLMs. While RGF produces a question closely related to the input, LLMs generate more diverse questions with respect to surface form and semantic content.}
\label{fig:cf_example}
\end{figure}

To alleviate the issue of poor OOD performance for QA, recent works have resorted to augmenting training data with \textit{counterfactual} instances automatically generated by LLMs \cite{paranjape-etal-2022-retrieval}.
Training on CF augmented data reduces model reliance on spurious
features, which in turn improves generalizability~\cite{sen-etal-2021-counterfactually}. 
While \newcite{paranjape-etal-2022-retrieval} fine-tune a T5-based model to generate minimally different counterfactual instances with their Retrieve-Generate-Filter~(RGF) approach, we leverage a range of more powerful LLMs such as Flan-UL2~\cite{tay2022ul2} and LLaMA~\cite{DBLP:journals/corr/abs-2302-13971}. 
Owing to the extensive training of these LLMs on diverse data, coupled with their enhanced generative capabilities, we hypothesize they will produce counterfactual instances \textit{more diverse} with respect to their surface form and semantic content, covering a broader part of the input space, further improving robustness and generalization. 
A sample of diverse CF instances is shown in \Cref{fig:cf_example}, highlighting variations in focus, temporality, specificity, and domain knowledge. 

In other work, \citet{ye-durrett-2022-explanations} improve the calibration of SLMs by leveraging features from \textit{rationales}, explanations of the inner decision making process of the model, to train a calibrator model -- a simple classifier which predicts whether the base model is correct or not.
We hypothesize that CF augmented models possess more precise explanations of their decisions, as they are forced to consolidate the more complex discrepancies between instances and their CFs, which should in turn provide better information to the calibrator model and improve calibration.
To better investigate the connection between model explanations and calibrator performance, we introduce semantic features -- dense representations of the most important tokens from explanations -- to calibrator models, consider a wider range of explainability methods, and measure whether characteristics of explanations -- such as \textit{comprehensiveness} and \textit{sufficiency}~\cite{chrysostomou-aletras-2022-empirical} are indicative of the models' calibration performance.

In our work, we present the first systematic and comprehensive study on the effect of diverse CFs for augmenting SLMs with respect to their OOD performance, explanation quality and calibration performance. Our experiments show that: (\underline{1}) more diverse CFs improve OOD performance and model calibration in extractive QA by a large margin; (\underline{2}) introducing rationale semantics from CF augmented models to calibrators improves calibration performance; and (\underline{3}) rationale augmented calibrators prefer concise and informative explanations.


\section{Related Work}

\subsection{Counterfactual Generation}

Counterfactual instances have demonstrated their importance in evaluating the OOD generalization capabilities of LLMs~\cite{bowman-dahl-2021-will} and in augmenting training data \cite{longpre-etal-2021-entity}. One major downside of works which tackle CF generation~\cite{Kaushik2020Learning,khashabi-etal-2020-bang,ribeiro-etal-2020-beyond} has been the prohibitive requirement for human annotators, which would manually perturb data instances to generate CFs -- a setup both expensive and difficult to scale.

With the improvements brought forward by LLMs, the idea of automatically generating CFs with generative models has gained significant traction. %
In the QA setup, \citet{ye-etal-2021-connecting} and \citet{longpre-etal-2021-entity} generate counterfactuals by substituting entity names with other plausible entity names. However, this approach requires heuristic methods or human re-labeling to derive the resulting label changes.
More recent work \citep{paranjape-etal-2022-retrieval} focuses on creating fluent, and automatically labeled CFs with minimal human supervision.
Their method requires fine-tuning models for both question generation and answering, which restricts the diversity of generated CFs to only what exists within the fine-tuning dataset.
On the other hand, our methodology utilizes LLMs pretrained on a diverse array of datasets that enables us to generate CFs with a broader range of knowledge and linguistic nuances, surpassing the limitations posed by fine-tuning on specific datasets. 
\citet{DBLP:journals/corr/abs-2310-00603} prompt LLMs to generate CFs by altering a specific attribute conveyed in the input text while confounding attributes are fixed. In contrast, our work emphasizes on generating diverse CF instances without the constraint of changing a specific input attribute.
In summary, our work investigates the previously unexplored relationship between CF diversity and OOD performance.

\subsection{Model Calibration}

Estimating the uncertainty of SLMs is challenging due to limited training data available, especially under OOD settings~\cite{desai-durrett-2020-calibration, guo-etal-2021-overview}. While prior approaches to model calibration have used ``meta-features'' based on model confidence~\citep{DBLP:conf/acl/KamathJL20} and input representations~\citep{zhang-etal-2021-knowing}, these techniques do not incorporate features from explanations which is the central focus of our work. 
In the OOD calibration scenario, recent works have explored the use of explanations during training~\citep{DBLP:conf/emnlp/LiHC22}, and data augmentation~\citep{park-caragea-2022-calibration}. However, these works mostly focus on calibration techniques, whereas token importance scores from explanations are only used for selecting data samples that improve model generalization.
More recently, \citet{ye-durrett-2022-explanations} studied how to improve a black box model’s calibration in OOD settings by leveraging handcrafted features from explanations~\cite{ribeiro-lime, lundberg-shap}. %
However, their method of computing handcrafted features maps tokens to linguistic features such as POS tags, a process in which the meaning of individual tokens is lost. 
In our work, we explore the connection between explanation content of CF augmented models and calibration performance. The questions we set out to answer are: (1) does the content of the explanation matter to the calibrator? (2) which explainer is the best at producing calibration features? and (3) which characteristics of explanations are important for calibration?

\section{Methodology}
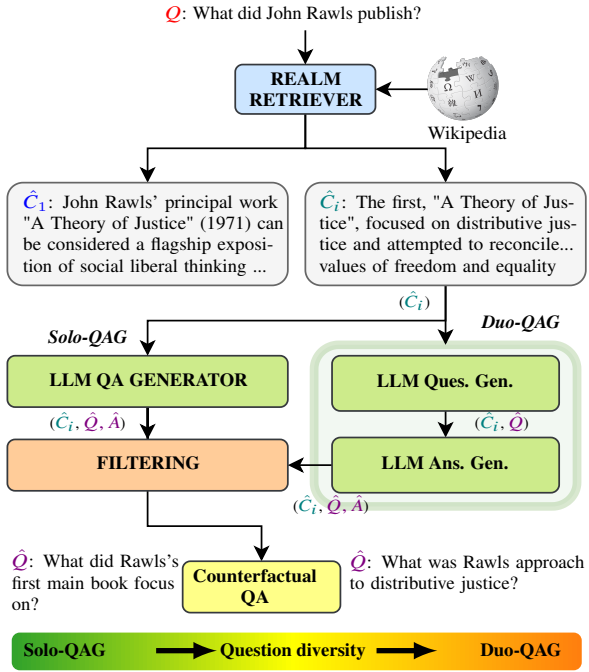
\begin{figure}[!t]
  \center
  \vspace{8pt}
  \includegraphics[width=0.48\textwidth, height=8cm]{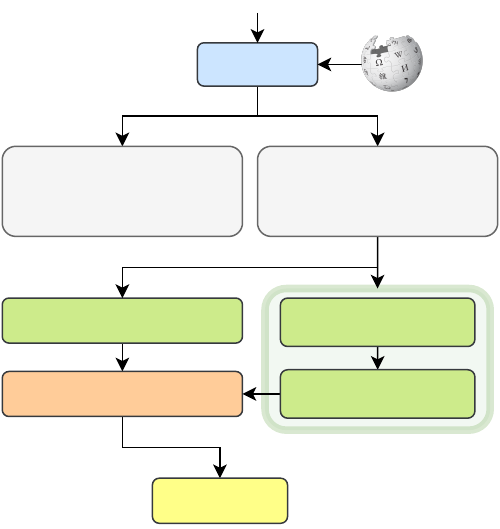}
    \begin{tikzpicture}[overlay, remember picture]

  \pgfdeclareverticalshading{myshading}{100bp}{
        color(0bp)=(red);
        color(25bp)=(orange);
        color(50bp)=(yellow);
        color(75bp)=(green);
        color(100bp)=(blue)
    }
    \shade[shading=myshading, rounded corners=3pt, shading angle=90] (-3.7,-0.2) rectangle (3.8,0.3);

    \node[text=black, font=\scriptsize, align=left] at (0, 0.05)
    {\textbf{Question diversity}};
    \draw[-latex, line width=2pt] (-1.8,0.06) -- (-1,0.06);
    \draw[-latex, line width=2pt] (1.1,0.06) -- (1.9,0.06);

    \node[text=black, font=\scriptsize, align=left] at (-3, 0.05)
    {\textbf{Solo-QAG}};
    \node[text=black, font=\scriptsize, align=left] at (3, 0.05)
    {\textbf{Duo-QAG}};

    \node[text=black, font=\scriptsize, align=left] at (0, 8.5)
    {\textcolor{red}{$Q$}: What did John Rawls publish?};

    \node[text=black, font=\scriptsize, align=center, text width=2cm] at (0.17, 7.5)
    {\textbf{REALM\\ RETRIEVER}};

    \node[text=black, font=\scriptsize, align=center, text width=2cm] at (2.3,6.9)
    {Wikipedia};

    \node[text=black, font=\scriptsize, align=left,  text width=3.5cm] at (-1.8, 5.6)
    {\textcolor{blue}{$\hat{C}_1$}: John Rawls' principal work "A Theory of Justice" (1971) can be considered a flagship exposition of social liberal thinking   ...};

    \node[text=black, font=\scriptsize, align=left,  text width=3.5cm] at (2.1, 5.6)
    {\textcolor{teal}{$\hat{C}_i$}: The first, "A Theory of Justice", focused on distributive justice and attempted to reconcile... values of freedom and equality};

    \node[text=black, font=\scriptsize, align=center] at (-2.7, 4.2) {\textit{\textbf{Solo-QAG}}};
    \node[text=black, font=\scriptsize, align=center] at (3, 4.4) {\textit{\textbf{Duo-QAG}}};
    
    \node[text=black, font=\scriptsize, align=center] at (-1.9, 3.63) {\textbf{LLM QA GENERATOR}};
    \node[text=black, font=\tiny, align=center] at (1.6, 4.7)
     {(\textcolor{teal}{$\hat{C}_i$})};
    \node[text=black, font=\tiny, align=center] at (2.75, 3.1)
     {(\textcolor{teal}{$\hat{C}_i$}, \textcolor{violet}{$\hat{Q}$})};
    \node[text=black, font=\tiny, align=center] at (-2.7, 3.1)
     {(\textcolor{teal}{$\hat{C}_i$}, \textcolor{violet}{$\hat{Q}, \hat{A}$})};
     \node[text=black, font=\tiny, align=center] at (0.5, 2)
     {(\textcolor{teal}{$\hat{C}_i$}, \textcolor{violet}{$\hat{Q}, \hat{A}$})};
    \node[text=black, font=\scriptsize, align=center] at (-1.9, 2.55) {\textbf{FILTERING}};
    \node[text=black, font=\scriptsize, align=center] at (2, 3.62) {\textbf{LLM Ques. Gen.}};
    \node[text=black, font=\scriptsize, align=center] at (2, 2.55) {\textbf{LLM Ans. Gen.}};
      \node[text=black, font=\scriptsize, align=left, text width=3.5cm] at (-1.95, 1)
    {\textcolor{violet}{$\hat{Q}$}: What did Rawls's \\first main book focus\\ on?};
    
    \node[text=black, font=\scriptsize, align=left, text width=3.5cm] at (2.5, 1.11)
    {\textcolor{violet}{$\hat{Q}$}: What was Rawls approach \\to distributive justice?};

    \node[text=black, font=\scriptsize, align=center, text width=2cm] at (-0.5, 0.9) {\textbf{Counterfactual\\ QA}};

  \end{tikzpicture}
    \vspace{0.8cm}
  \label{ref:fig:method}
  \caption{Our proposed methodology for generating counterfactual instances. The Solo-QAG approach~(left) generates counterfactual QA pairs in a single pass while the Duo-QAG approach~(right) first generates the question, and then the answer. }
  \end{figure}

  \begin{figure}[!t]
  \center
  \includegraphics[width=0.48\textwidth, height=6cm]{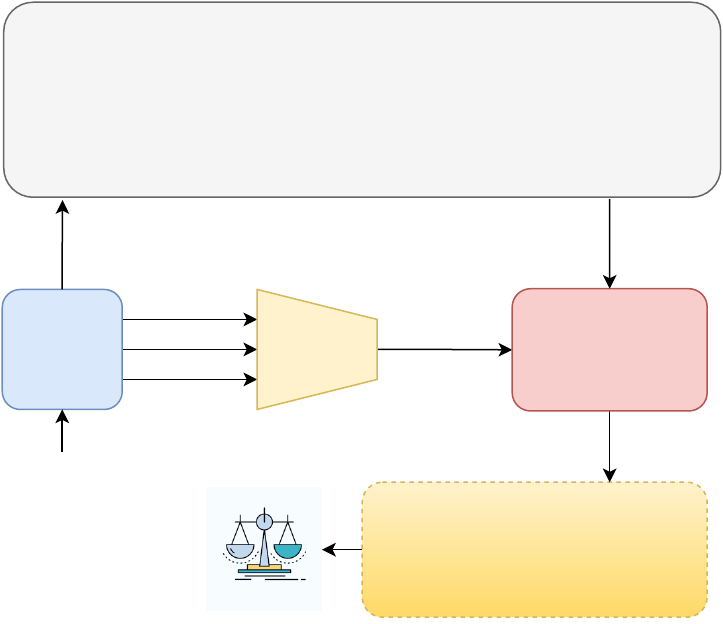}
    \begin{tikzpicture}[overlay, remember picture]

    \fill[yellow] (-2.4,6) rectangle (-1.8,6.3);
    \fill[yellow!70] (-1.7,6) rectangle (-0.5,6.3);
    \fill[yellow!50] (-0.44,6) rectangle (-0.25,6.3);
    \fill[yellow!30] (-1.2,5.7) rectangle (-0.95,6);

    \fill[yellow] (0,4.95) rectangle (0.5,5.25);
    \fill[yellow!70] (-3.6,4.7) rectangle (-2.9,5);
    \fill[yellow!50] (-1.8,4.7) rectangle (-1.2,5);
    
    \node[text=black, font=\scriptsize, align=left, text width=7.5cm] at (0.28,6)
    {\textbf{Question}: What anniversary of the Pokémon franchise was celebrated during the Super Bowl?};

    \node[text=black, font=\scriptsize, align=left, text width=7cm] at (0,5.1)
    {\textbf{Context}: Nintendo and The Pokémon Company also made their Super Bowl debut, promoting the \textbf{20th} anniversary of the Pokémon video game and media franchise.};

    \node[text=black, font=\scriptsize, align=center, text width=2cm] at (-3.15,3.15)
    {\textbf{CF\\ augmented\\ model}};

    \node[text=black, font=\scriptsize, align=center, text width=2cm] at (-3.1,1.2)
    {Input\\ question\\ $+$\\ context};

     \node[text=black, font=\scriptsize, align=center, text width=2cm, rotate=90] at (-3.4,4.1)
    {SHAP};

     \node[text=black, font=\scriptsize, align=center, text width=2cm] at (-1.85,3.4)
    {Last hidden\\ states};

    \node[text=black, font=\scriptsize, align=center, text width=2cm] at (-1.8,2.3)
    {$(1, 512, 768)$};

    \node[text=black, font=\scriptsize, align=center, text width=2cm] at (-0.5,3.1)
    {\textbf{PCA}};

     \node[text=black, font=\scriptsize, align=center, text width=2cm] at (0.95,3.12)
    {Reduced\\ states};

    \node[text=black, font=\scriptsize, align=center, text width=2cm] at (0.95,2.7)
    {$(1, 512, 10)$};

    \node[text=black, font=\scriptsize, align=center, text width=2.7cm] at (1.7, 4.15)
    {Topk\\ (Context, Answer)\\ tokens};

    \node[text=black, font=\scriptsize, align=center, text width=2cm] at (2.65,3.4)
    {\textbf{ State Selector}};

     \node[text=black, font=\scriptsize, align=center, text width=2cm] at (2.7,3.05)
    {$(1, \text{topk}_C, 10)$};

    \node[text=black, font=\scriptsize, align=center, text width=2cm] at (2.7,2.8)
    {$(1, \text{topk}_A, 10)$};

    \node[text=black, font=\scriptsize, align=center, text width=2cm] at (3.1,2.2)
    {states};

    \node[text=black, font=\scriptsize, align=center, text width=2cm] at (1.8,1.6)
    {\textbf{ Featurize}};

    \node[text=black, font=\scriptsize, align=center, text width=2cm] at (1.1,1.1)
    {$\frac{1}{\text{topk} (C,A)} \sum_{j=0}^{\text{topk} (C,A)} \text{states}_{(1, j, 10)}$};

    \node[text=black, font=\scriptsize, align=center, text width=2cm] at (-1,0.1)
    {Prediction: \textcolor{green}{correct} or \textcolor{red}{incorrect}};

  \end{tikzpicture}
    \vspace{0.5cm}
  \caption{Our proposed calibration methodology. The dense representations of the highly important input tokens from the CF-augmented model are condensed and converted to semantic features to train a classifier that predicts if the model prediction is correct.}
  \label{ref:fig:calib_method}
  \end{figure}
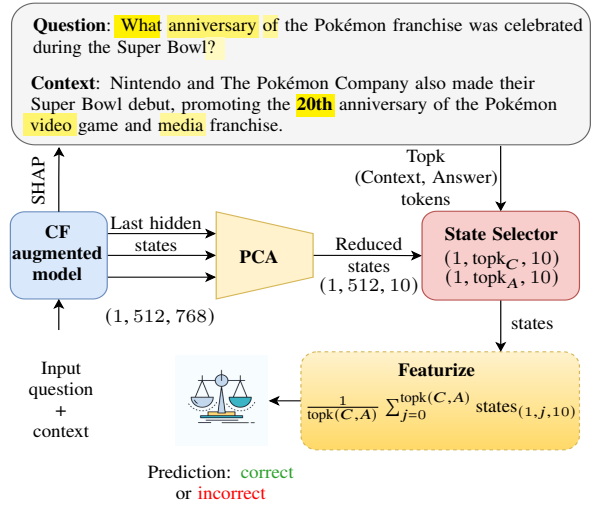

\subsection{Datasets} \label{sec:meth_datasets}

We evaluate our CF augmentation methods on seven extractive question answering datasets commonly used in related works: SQuAD~\cite{rajpurkar-etal-2016-squad}, SQuAD-Adversarial~\cite{DBLP:conf/emnlp/JiaL17}, TriviaQA~\cite{DBLP:conf/acl/JoshiCWZ17}, HotpotQA~\cite{DBLP:conf/emnlp/Yang0ZBCSM18}, Natural Questions~(NQ)~\cite{kwiatkowski-etal-2019-natural}, NewsQA~\cite{trischler-etal-2017-newsqa}, BioASQ~\cite{DBLP:journals/bmcbi/TsatsaronisBMPZ15}. For all datasets except SQuAD, we directly use the pre-processed version of the dataset from the MRQA Shared Task~\cite{DBLP:conf/acl-mrqa/FischTJSCC19}. We provide detailed descriptions of the datasets in \Cref{app:datasets}.

\subsection{Setup and Base Models} \label{sec: meth_setup}
Following the setup of \citet{ye-durrett-2022-explanations}, we train a RoBERTa-base model~\cite{DBLP:journals/corr/abs-1907-11692} on the SQuAD dataset and evaluate its OOD performance on the remaining six datasets. 
To improve the generalization capabilities of our base model, we augment the SQuAD data with CFs automatically generated using the following LLMs: (\underline{1}) GPT-JT~(6B) and (\underline{2}) GPT-NeoxT~(20B), instruction tuned versions of GPT-J~\cite{gpt-j} and GPT-Neox~\cite{DBLP:journals/corr/abs-2204-06745}; (\underline{3}) LLaMA~(13B)~\cite{DBLP:journals/corr/abs-2302-13971}, (\underline{4}) Alpaca~\cite{alpaca}, (\underline{5}) Flan-T5-xxl~(11B)~\cite{DBLP:conf/iclr/WeiBZGYLDDL22}, and (\underline{6}) Flan-UL2~(20B)~\cite{tay2022ul2}. 
We obtain the Alpaca model by Low-Rank Adaptation~(LoRA)~\cite{hu2022lora} fine-tuning the LLaMA~(13B) model on the Alpaca dataset~\cite{alpaca} for 10 epochs.
These models are selected as they are publicly available, trained on varying data and representative of both decoder-only and encoder-decoder families, as well as their instruction-tuned variants.
We omit detailed model descriptions for brevity and refer the reader to \Cref{app:models} for more details.

\subsection{Generating Counterfactuals}

\subsubsection{Retrieve-Generate-Filter} \label{sec:method_rgf}

Introduced in \citet{paranjape-etal-2022-retrieval}, retrieve-generate-filter~(RGF) describes a framework used to create counterfactual instances with minimal human supervision.
The \textit{retrieval} step leverages the REALM retrieval augmented language model~\cite{10.5555/3524938.3525306} to produce a ranked list of contexts and answers within those contexts, given a question as input.
Based on this set of contexts and answers, RGF then \textit{generates} question candidates using a T5-3B
question generation model fine-tuned on the NQ dataset. These question candidates are then \textit{filtered} to ensure quality.
For the sake of space, we elaborate the details of all filtering steps in \Cref{app:filtering_cfs}.
Each generated question, along with its corresponding context and answer, constitutes a \textit{counterfactual} instance. %



\subsubsection{Solo-QAG and Duo-QAG}
In our proposed approaches, we first use the REALM model to \textit{retrieve} candidate contexts and then select the context for which the T5-large model generates the closest question based on the Levenshtein distance~\cite{Levenshtein_SPD66}. 
Then, given the chosen context $\hat{c}_i$, our \textit{LLM QA generator} generates the counterfactual $(\hat{q},\hat{a})$ pair using an LLM prompted in $1$-shot manner with a prompt containing the original $(q, a)$ pair.
As our preliminary experiments have shown that some LLMs are better at jointly generating the question and answer, while others perform better at sequential generation, we propose the following two approaches to generating counterfactual instances: %

\paragraph{Solo-QAG.} For every chosen context $\hat{c}_i$, we prompt the LLM to produce a question-answer pair ($\hat{q}_i$, $\hat{a}_i$) in a single generative step. We name this approach \textbf{S}ingle-\textbf{P}hase \textbf{Q}uestion-\textbf{A}nswer \textbf{G}eneration.

\paragraph{Duo-QAG.} In this approach, we split LLM QA generation into two phases. We first generate the question $\hat{q}_i$ that can be answered based on the given context $\hat{c}_i$, and then use the question-context pair ($\hat{q}_i$, $\hat{c}_i$) to generate an answer $\hat{a}_i$. We name this approach \textbf{D}ual-Phase \textbf{Q}uestion-\textbf{A}nswer \textbf{G}eneration.

\paragraph \noindent We illustrate our proposed approaches in \Cref{ref:fig:method}. In both generation approaches, we prompt the LLM a maximum of three times with different random seeds until a satisfiable instance is produced (e.g. one which is not empty or excessively short).
We detail the prompts used for CF generation in \Cref{app:prompts}.

As the LLM-based CF generation approaches are still prone to generating open-ended questions which cannot be answered based on information provided in the input context, we introduce a filtering step designed to ensure high quality of generated CF instances.
The first filtering step leverages \textit{context relevance filtering} to identify CF questions where the corresponding input context does not provide sufficient information for an answer. 
Since context relevance filtering may also discard some complex, but answerable questions, we further employ the round-trip consistency approach~\cite{alberti-etal-2019-synthetic, DBLP:journals/corr/abs-2009-05167} to retrieve incorrectly discarded samples using an ensemble of three language models initialized with different seeds to answer the LLM generated questions. If answers from 2 or more language models agree with the LLM-generated answer, the CF sample is retained.  

\paragraph{Intrinsic Evaluation.}
We evaluate the generated counterfactuals along two dimensions: \textit{Fluency} and \textit{Correctness}. Fluency measures whether the generated CF question is grammatically correct and semantically meaningful. Correctness measures the alignment between the generated question $\hat{Q}$, context $\hat{C}$, and answer $\hat{A}$, i.e., the question is answerable from the context and the answer is correct.

\begin{figure}[!t]
  \center
  \includegraphics[width=0.48\textwidth]{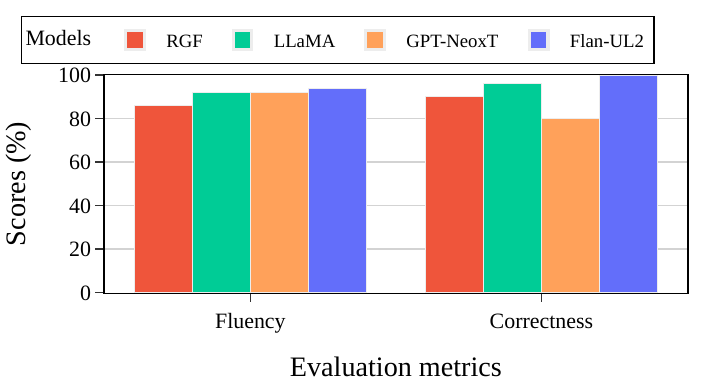}
  \caption{Quantitative evaluation of fluency and correctness of the CF instances generated by the RGF, LLaMA, GPT-NeoxT, and Flan-UL2 models.}
  \label{fig: cf_evaluation}
  \end{figure}

We perform a human evaluation on 50 CF instances sampled each from the RGF, LLaMA, GPT-NeoxT, and Flan-UL2 models and report our results in \Cref{fig: cf_evaluation}. We find that over 90\% of the generated questions from all the models are fluent, as the generation leverages high-quality pre-trained language models.
We further quantify the correctness of the generated CF instances and find out that our methodology with LLaMA and Flan-UL2 models produces minimal~(\textless5\%) noisy data as compared to \textasciitilde{20}\% of RGF, stating that the Solo-QAG and Duo-QAG produce superior and answerable CF instances. We detail the annotation process in \Cref{app:human_eval}.

\subsubsection{Estimating Diversity of Counterfactuals}
\label{sec: cf_eval}
We quantitatively evaluate the \textit{diversity} of generated counterfactual questions with respect to the original questions along two axes: (\underline{1}) \textit{surface form variation}, measured by self-BLEU~\cite{10.1145/3209978.3210080} and Levenshtein edit distance, as proposed in \citet{wu-etal-2021-polyjuice}; and (\underline{2}) \textit{semantic variation}, measured by SBERT~\cite{DBLP:conf/emnlp/ReimersG19} embedding similarity and semantic uncertainty~\cite{DBLP:conf/iclr/KuhnGF23}. 
\textit{Surface form variation} metrics quantify the surface form difference between the original question and its counterfactual counterpart through n-gram and character level overlaps.
Lower self-BLEU and conversely, a higher edit distance indicate greater surface form diversity between a question and its corresponding CF.
As surface form diversity does not necessarily imply semantic difference, we also estimate \textit{semantic variation} through two methods.
Our first method estimates semantic diversity through cosine similarity between the SBERT embeddings of a question and its counterfactual counterpart.
We complement SBERT similarity by adapting a novel method measuring semantic uncertainty. 
Following \citet{DBLP:conf/iclr/KuhnGF23}, we leverage a pretrained natural language inference model, in our case DeBERTa-large~\cite{DBLP:conf/iclr/HeLGC21} and compute the bidirectional entailment (equivalence) probability between the original question and its corresponding CF.
Herein, a lower equivalence score indicates lower confidence of entailment between the pair, which in turn corresponds to greater semantic variation.
In Appendix \ref{app: gen_cf},  we highlight semantic variations introduced in randomly sampled CFs generated by our approach.

\subsection{Model Calibration}
As prediction probabilities of SLMs are often poorly calibrated, practitioners frequently resort to model calibration -- training simpler models to detect when the underlying model is faulty by producing a score which overrides the model confidence and conveys whether the original prediction is correct.
The benefit of a good calibrator model is that faulty, but confident, predictions can be detected before a wrong answer is returned to an end-user.
Apart from the base model confidence, these models usually leverage diverse heuristic features as additional inputs to a calibrator model.
Following previous work \cite{DBLP:conf/acl/KamathJL20, ye-durrett-2022-explanations}, we use the random forest classifier as our calibration model. 
We train each calibrator model on $500$ training samples, changing only the input feature sets, while the correctness of the base model prediction is used as the output label.
To evaluate the quality of model calibration, we leverage the Macro-average Calibration Error, MacroCE \citep{si-etal-2022-examining}, a recently proposed enhanced version of the Expected Calibration Error (ECE) \citep{10.5555/3305381.3305518}. 
We elaborate the model calibration procedure in~\Cref{app:calib_details}.

\subsubsection{Baseline} \label{sec:meth_calib_baseline}

\citet{ye-durrett-2022-explanations} focus on calibrating black-box models with explainers based on local perturbation techniques: LIME~\cite{ribeiro-lime} and SHAP~\cite{lundberg-shap}. Due to the large scale of our experiments and the high computational complexity of LIME, we only use their SHAP feature-based calibration technique.
\citet{ye-durrett-2022-explanations} map input tokens to linguistic features, such as POS tags, and then aggregate importance scores across all tokens assigned specific feature values, e.g. nouns.
These aggregated scores are used as input features of the calibrator model, augmenting it with the explanation information. 

\subsubsection{Improving Explanations for Calibrators}\label{sec:our_calib}

The calibration approach of \citet{ye-durrett-2022-explanations} has two main limitations: 
1) they consider only black-box explainability methods, leaving uncertainty about calibrators' preferences for alternative explainability methods; and
(2) input features to calibrator models are aggregated importance values of tokens from specific word categories (i.e. POS tags), a process where the token meanings are lost.

To tackle the first issue and account for the variation in quality of explanations generated by explainability methods~\cite{jain-wallace-2019-attention, DBLP:conf/hhai/NeelySBL22}, we drop the restrictive black-box scenario and extend the scope of our evaluation to attention- and gradient-based white-box explainers, which provide a broader overview of how explanations affect calibration performance. We employ normalized attention scores~($\alpha$)~\cite{jain-etal-2020-learning} and gradient-scaled attention scores~($\alpha\nabla\alpha$) from the attention-based family, while we consider InputXGradients~($x\nabla x$)~\cite{DBLP:journals/corr/KindermansSMD16} and integrated gradients (IG)~\citep{10.5555/3305890.3306024} from gradient-based approaches.
To address the second drawback, we augment calibrator models with semantic features computed from dense representations of input tokens assigned high importance by explanation methods. %
We select the top $10$\% and $20$\% most salient tokens from the \textit{context} and \textit{answer}, respectively, reduce their dimensionality to the top ten principal components
 using PCA~\cite{DBLP:journals/corr/Shlens14}\footnote{The number of principal components was determined through a series of non-exhaustive experiments. We experimented with $10$ and $100$ features, and found that using $10$ features yields better results.}, and then average their token representations. The resulting vector is then used as additional input to the calibrator model. 
We sketch our proposed calibration procedure in \Cref{ref:fig:calib_method}.

We are also interested in the explanation characteristics indicative of the rationale-augmented calibrators' performance.
To this end, we measure the \textit{comprehensiveness} and \textit{sufficiency}~\cite{DBLP:conf/acl/DeYoungJRLXSW20} of generated explanations, two metrics used to determine the influence of the rationale on a prediction.
Given input tokens $\{x_i\}_{i=1}^t$, \textit{comprehensiveness} masks $n\%$ input tokens assigned the highest importance scores. 
The comprehensiveness score is then determined as the change in the prediction probability of the model for the same answer, where a high difference in the prediction score indicates that the masked rationale tokens were influential for the prediction.
To estimate the degree to which extracted rationales are \textit{sufficient} for the models' prediction, given input tokens $\{x_i\}_{i=1}^t$, \textit{sufficiency} retains only $n\%$ of tokens assigned the highest importance scores, masking out the rest. The sufficiency score is then determined as the change in prediction probability of the model for the same answer.
Following~\citet{carton-etal-2020-evaluating} and~\citet{chrysostomou-aletras-2022-empirical}, we constrain sufficiency between $0$ and $1$ and report \textit{$1-\text{suff}$} so that higher is better. %
        
In case of extractive QA, we do not mask~(for comprehensiveness) and explicitly keep~(for sufficiency) the question and answer tokens so that the model is able to answer the input question. We report average sufficiency and comprehensiveness scores when retaining~(for sufficiency) or masking~(for comprehensiveness) the top $n \in \{2\%, 10\%, 20\%, 50\%\}$ most important tokens.

\section{Experiments}

\renewcommand{\arraystretch}{0.8}
\begin{table*}[!t]
\small
\centering
\begin{tabularx}{\textwidth}{@{}p{2cm} p{2cm} ccccc@{}}
\toprule

\multirow{2}{*}{\textbf{Approach}} & \multirow{2}{*}{\textbf{Model}} & \multicolumn{2}{c}{\textbf{Surface form variation}} & \multicolumn{2}{c}{\textbf{Semantic variation}} \\
\cmidrule(lr){3-4} \cmidrule(lr){5-6} 
&& Self-BLEU ($\downarrow$) & Levenshtein ($\uparrow$) & SBERT Sim. ($\downarrow$)  & Semantic Equivalence ($\downarrow$) \\
\midrule
Reference & -  & 0.11 & 1.00 & 0.11  & 0.54 \\
\midrule
 \textsc{RGF} & \textsc{T5-3B}  & 0.31 & 0.61 & 0.56  & 0.52 \\
\midrule
\multirow{4}{*}{\textsc{Solo-QAG}}& GPT-JT & 0.26 & 0.67 & 0.48  & 0.46 \\
& \textsc{LLaMA}  & 0.28 &  0.65 & 0.50  & 0.51\\
& \textsc{Alpaca}   & 0.27 & 0.67 & 0.50  & 0.55 \\
& \textsc{GPT-NeoxT}  & 0.24 & 0.68 & 0.45  & 0.46 \\
\midrule
\multirow{2}{*}{\textsc{Duo-QAG}}& \textsc{Flan T5-xxl}  & \textbf{0.19} & \textbf{0.71} & \textbf{0.41}  & 0.41\\
& \textsc{Flan-UL2}  & \textbf{0.19} & \textbf{0.71} & \textbf{0.41}  & \textbf{0.40} \\
\bottomrule
\end{tabularx}
\caption{Quantitative evaluation of the diversity of generated counterfactuals with respect to the original questions. The metrics are complementary -- diverse CFs are expected to be further away from original instances in both surface form and meaning. To contextualize semantic and surface form variation of CFs, we contrast them to a \textbf{reference} baseline -- diversity of an instance compared to a randomly selected other instance from the dataset.}
\label{table:diversity}
\end{table*}

\subsection{Generating Counterfactual Instances}

We report an overview of models used to generate CFs, their parameter sizes and the resulting number of generated (usable) CFs in  \Cref{app: gen_cf}.
The Duo-QAG approach yields a significantly higher number of usable samples~(\textasciitilde70k) compared to Solo-QAG~(\textasciitilde50k), indicating that the two-step approach produces higher fidelity CF instances.
We hypothesize that the better generative abilities of the Duo-QAG approach arise from the extensive pre-training of FLAN-based LLMs on question generation and question-answering tasks.

In \Cref{table:diversity}, we report the diversity of the generated CFs with respect to surface form and semantic variation. 
Our reference approach quantifies the upper bound of the SQuAD dataset diversity by comparing every data sample with another random sample from the dataset. 
The RGF approach produces the least diverse CFs, which is expected considering its methodology which aims to generate and select CFs which deviate minimally from the input samples. 
Contrary to RGF, our methodology utilizes capabilities of LLMs to produce CF instances that are semantically and contextually more diverse. 
We hypothesize that counterfactual instances more diverse from the original improve the models' input space coverage, which should in turn improve OOD performance and calibration. 
We verify this hypothesis in the following sections.

\begin{table*}[!t]
\small
\centering
\begin{tabularx}{\textwidth}{l| *{7}{>{\centering\arraybackslash}X}| >{\centering\arraybackslash}p{0.9cm}}
\toprule
\textbf{Exact Match} & \textbf{SQuAD} & \textbf{SQuAD$_{Adv.}$} & \textbf{TriviaQA} & \textbf{HotpotQA} & \textbf{NQ} & \textbf{NewsQA} & \textbf{BioASQ}  & \textbf{$G_{\text{ood}}$} \\

\midrule
\textsc{Base}
& $84.98_{\mbox{\tiny 0.07}}$  & $66.60_{\mbox{\tiny0.84}}$  & $39.09_{\mbox{\tiny1.87}}$ & $48.16_{\mbox{\tiny0.16}}$ &  $41.94_{\mbox{\tiny1.01}}$ & $42.21_{\mbox{\tiny0.69}}$ & $47.93_{\mbox{\tiny1.22}}$ & -\\
 
 \midrule

 \textsc{RGF}
& \cellcolor{green!20}\textbf{$\mbox{85.53}_{\mbox{\tiny0.04}}$}  & $65.97_{\mbox{\tiny0.42}}$  & $44.98_{\mbox{\tiny0.22}}$ & $52.88_{\mbox{\tiny0.25}}$ &  $46.22_{\mbox{\tiny0.29}}$ & \cellcolor{green!20}\textbf{$\mbox{43.01}_{\mbox{\tiny0.32}}$} &  $50.50_{\mbox{\tiny0.23}}$ & $2.94$\\
 
 \midrule

\textsc{GPT-JT}  & $84.74_{\mbox{\tiny0.18}}$  & $67.19_{\mbox{\tiny0.42}}$  & \cellcolor{orange!20}$47.40_{\mbox{\tiny0.33}}$  & $51.21_{\mbox{\tiny0.50}}$  & $47.08_{\mbox{\tiny0.84}}$  & $42.12_{\mbox{\tiny0.59}}$ & \cellcolor{orange!20}$52.59_{\mbox{\tiny1.13}}$  & $3.61$\\

\textsc{LLaMA}  & $84.85_{\mbox{\tiny0.31}}$  & $67.57_{\mbox{\tiny0.42}}$ & \cellcolor{green!20}\textbf{$\mbox{48.13}_{\mbox{\tiny0.05}}$}  & $51.68_{\mbox{\tiny0.86}}$ & \cellcolor{orange!20}$48.80_{\mbox{\tiny0.68}}$ & $42.35_{\mbox{\tiny0.47}}$ & $51.68_{\mbox{\tiny1.25}}$ & $4.05$ \\

\textsc{Alpaca}  & \cellcolor{orange!20}$85.42_{\mbox{\tiny0.23}}$  & $66.59_{\mbox{\tiny0.98}}$  & $41.79_{\mbox{\tiny1.35}}$  & $51.88_{\mbox{\tiny0.55}}$  & $44.79_{\mbox{\tiny2.22}}$ & $42.48_{\mbox{\tiny0.55}}$ & $49.56_{\mbox{\tiny0.50}}$ & $1.86$ \\

\textsc{GPT-NeoxT}  & $84.80_{\mbox{\tiny0.25}}$  & \cellcolor{orange!20}$68.07_{\mbox{\tiny1.30}}$  & $46.96_{\mbox{\tiny0.46}}$ & $53.14_{\mbox{\tiny0.67}}$  & $47.80_{\mbox{\tiny1.83}}$ & $41.99_{\mbox{\tiny0.73}}$ & \cellcolor{green!20}\textbf{$\mbox{53.19}_{\mbox{\tiny1.06}}$}  & \cellcolor{green!20}\textbf{$\mbox{4.20}$}\\

\midrule

\textsc{Flan-T5-xxl} & $85.41_{\mbox{\tiny0.28}}$  & $67.15_{\mbox{\tiny0.59}}$ & $42.91_{\mbox{\tiny0.53}}$  & \cellcolor{orange!20}$53.52_{\mbox{\tiny0.95}}$  & $48.05_{\mbox{\tiny0.86}}$ & $42.70_{\mbox{\tiny1.18}}$ & $49.29_{\mbox{\tiny0.52}}$ & $2.95$ \\

\textsc{Flan-UL2}   & $85.38_{\mbox{\tiny0.10}}$ & \cellcolor{green!20}\textbf{$\mbox{68.09}_{\mbox{\tiny0.82}}$}  & $45.40_{\mbox{\tiny1.24}}$  & \cellcolor{green!20}\textbf{$\mbox{53.70}_{\mbox{\tiny0.56}}$}  & \cellcolor{green!20}\textbf{$\mbox{48.88}_{\mbox{\tiny0.91}}$} & \cellcolor{orange!20}$42.99_{\mbox{\tiny0.81}}$ & $51.33_{\mbox{\tiny0.48}}$ & \cellcolor{orange!20}$4.08$\\

\bottomrule
\end{tabularx}
\caption{EM results for RoBERTa-base model trained on the SQuAD dataset~(\textsc{Base}) and augmented with counterfactual data. We report the mean\textsubscript{std.} over 3 runs with different random seeds.
The last column ($G_{\text{ood}}$) shows the average gain over the \textsc{Base} model on OOD datasets. Numbers marked in \textbf{\textcolor{green}{bold green}}, and \textcolor{orange}{orange} colours represent the highest and second highest scores. 
We also report the F1 scores, which follow a similar trend, in \Cref{app:cf_gen_complete}.} 
\label{table:ood_results}
\end{table*}

\subsection{Generalization of CF Augmented Models}

We report the exact-match scores of the CF-augmented RoBERTa-base model on six OOD datasets in Table~\ref{table:ood_results}. 

Models augmented with CFs generated by our approach outperform all baselines across all OOD datasets, except NewsQA. 
We hypothesize that this is due to the complex reasoning required by NewsQA, involving synthesis of information from multiple sentences~\cite{trischler-etal-2017-newsqa} and that LLMs might not be able to generate diverse yet useful complex CF questions based on instances from the simpler SQuAD dataset.
All CF-augmented models maintain a comparable performance on the in-domain SQuAD dataset, implying that training with diverse data improves OOD generalization while preserving in-domain performance.
The spread of best-performing models shows that there is no \textit{one-model-fits-all} strategy and that even less diverse CFs may be better suited for some OOD datasets. 
The size and training data of the CF generator LLMs may also play an influential role as the larger scale LLaMA, GPT-NeoXT, and Flan-UL2 models are also the best performers. 
However, this aspect should not limit the applicability of our approach since even the smaller GPT-JT model provides significant gains on OOD datasets. 

Overall, the GPT-NeoxT CF augmented model has the highest average gain across all OOD datasets, with FLAN-UL2 and LLaMA closely behind. 
This is largely attributed to its strong performance on the BioASQ dataset, likely due to its pre-training on medical data from the large-scale PubMed Central dataset~\cite{DBLP:journals/corr/abs-2101-00027}. 
Our findings show that although all CF-augmented models consistently outperform baselines, the best augmentation approach depends on the concrete OOD dataset, suggesting that alignment between domain expertise of LLMs used to generate CFs and the data distribution of OOD datasets is important. 

\begin{figure*}[!t]
\begin{subfigure}{\textwidth}
    \includegraphics[width=\linewidth]{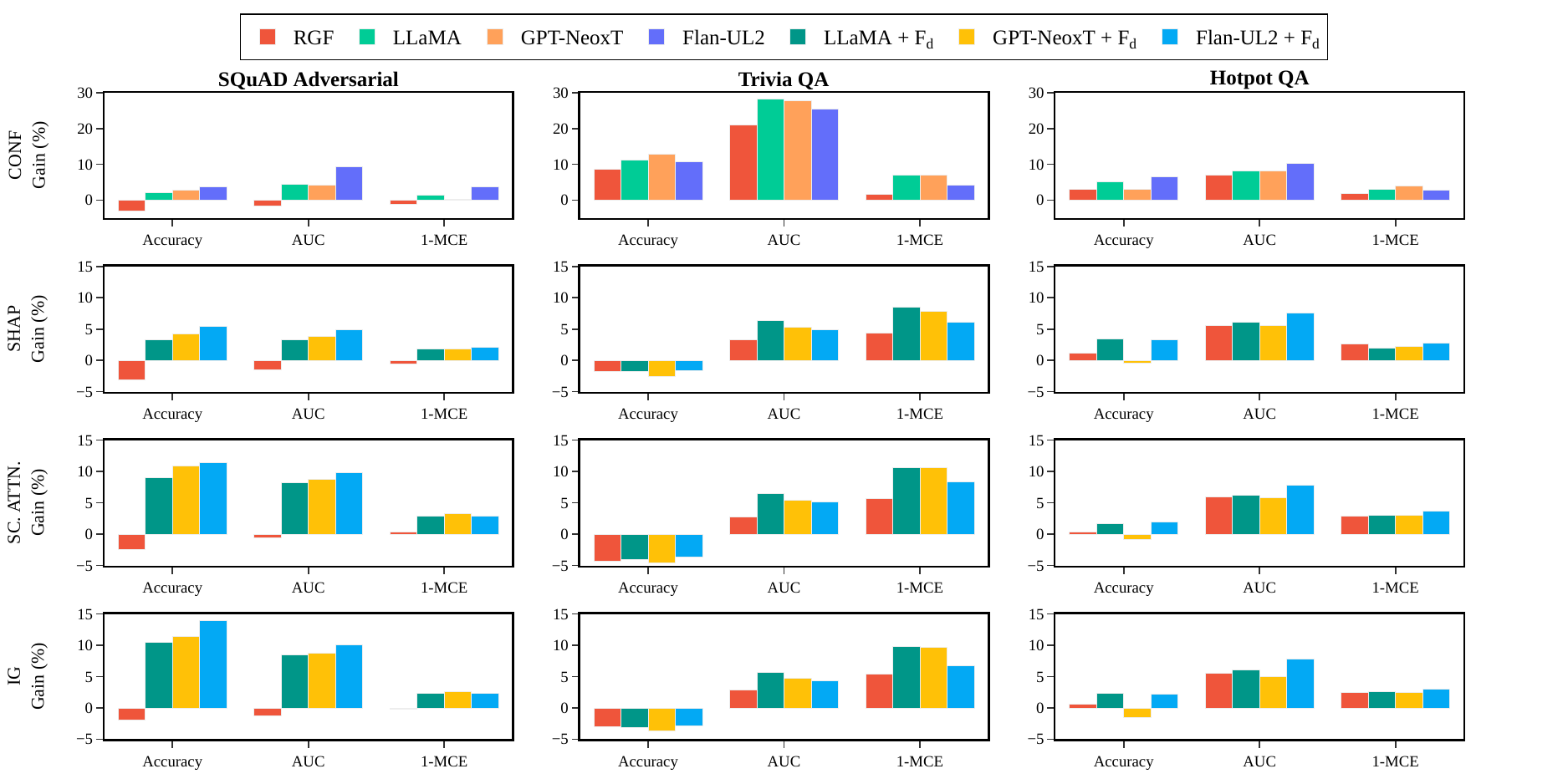}
\end{subfigure}\hfill

\caption{Percentage improvement of CF augmented models' calibration performance over the unaugmented RoBERTa-base model trained on SQuAD, using features based on probability~(\textsc{conf}) and rationales from \textsc{shap}, scaled attention and integrated gradients.
The results for \textsc{conf} (row \#1) are reported on models which do not use explanation-based features.
In the remaining experiments (other rows), along with \textsc{base} and \textsc{rgf}, we report the results of dense-feature augmented calibrators. We provide the complete results with other datasets and explanation methods in \Cref{app:model_calib}.} 
\label{fig:calib_results}
\end{figure*}

\subsection{Model Calibration}
We report the model calibration results as \% improvement over the base model in \Cref{fig:calib_results}.
We compare our models against two baselines: (1) \textsc{conf}, where the calibrator model only uses the thresholded probability of the predicted class to assess whether the prediction is trustworthy, and (2) \textsc{shap}. %
On the \textsc{conf} baseline, when only the probability of the underlying model is used as input to the calibrator, our CF-augmented models improve calibration accuracy across all OOD datasets with an average increase of \textasciitilde$5\%$, and up to \textasciitilde$11\%$ on the TriviaQA dataset.
These results suggest that augmenting a model with counterfactual instances already improves the model's capability to capture nuanced shifts in the data distribution. 
Improved robustness of CF-augmented models is further evident from the high inverse MacroCE scores on the \textsc{conf} baseline where even without features from explanations, CF-augmented models exhibit the best calibration scores~(\textasciitilde$+6\%$) across all datasets. 

When incorporating explanation features, on the \textsc{shap} baseline, the CF-augmented models improve calibration accuracy by an average of \textasciitilde$3\%$ on two out of three OOD datasets, the exception being TriviaQA, where the accuracy decreases marginally. 
Nevertheless, the CF-augmented models achieve superior AUC scores on all OOD datasets with an average improvement of \textasciitilde$5.5\%$ compared to the SHAP baseline without CF augmentation. 
For completeness, we report results on the NQ, NewsQA, and BioASQ datasets, along with the results produced by $\alpha$ and $x\nabla x$ in the \Cref{app:model_calib}. 

Overall, the CF-augmented models coupled with dense rationale features improve calibration over all baselines, all explanation methods, and OOD datasets, specifically on the SQuAD adversarial dataset. 
Our results show that augmenting training data with CF instances improves model calibration and that calibrators benefit from the semantic content of the most salient tokens from explanations. %


\begin{table*}[t]
\small
\centering
\setlength{\tabcolsep}{10pt}
\begin{tabularx}{\textwidth}{@{}lllllll|llllX}
\toprule
\multirow{2}[3]{*}{} & \multirow{2}[3]{*}{Model} & \multicolumn{5}{c}{\textbf{Comprehensiveness~($\uparrow$)}} & \multicolumn{5}{c}{\textbf{Sufficiency~($\uparrow$)}} \\
\cmidrule(lr){3-7} \cmidrule(lr){8-12}  
 && $\alpha$ & $\alpha\nabla\alpha$ & x$\nabla$x & IG & SHAP & $\alpha$ & $\alpha\nabla\alpha$ & x$\nabla$x & IG & SHAP\\
\midrule

& Base & 0.33 & 0.34 &  0.36 & 0.38 & 0.35 & 0.51 & 0.51 & 0.52 & 0.52 & 0.51 \\
& RGF & \cellcolor{green!20}\textbf{0.35} & \cellcolor{green!20}\textbf{0.41} &  \cellcolor{green!20}\textbf{0.41} & \cellcolor{green!20}\textbf{0.43} & \cellcolor{green!20}\textbf{0.44} & \cellcolor{red!20}0.41 & \cellcolor{red!20}0.43 & \cellcolor{red!20}0.42 & \cellcolor{red!20}0.43 & \cellcolor{red!20}0.41 \\
 & LLaMA & 0.32 & 0.34 & 0.33 & 0.34 & 0.32 & 0.54 & 0.55 & 0.54 & 0.54 & 0.54 \\
 & GPT-Neox &\cellcolor{red!20}0.29 & \cellcolor{red!20}0.31 &  \cellcolor{red!20}0.31 & \cellcolor{red!20}0.33 & \cellcolor{red!20}0.30 & \cellcolor{green!20}\textbf{0.56} & \cellcolor{green!20}\textbf{0.57} & \cellcolor{green!20}\textbf{0.57} & \cellcolor{green!20}\textbf{0.57} & \cellcolor{green!20}\textbf{0.56} \\
 & Flan-UL2 & 0.33 & 0.36 &  0.37 & 0.38 & 0.36 & 0.47 & 0.48 & 0.48 & 0.48 & 0.48 \\

\bottomrule
\end{tabularx}
\caption{Comprehensiveness and sufficiency scores of explanations generated by baseline and counterfactual augmented models, averaged across the six OOD datasets. 
Numbers marked in \textcolor{green}{\textbf{bold green}}, and \textcolor{red}{red} represent the highest and lowest scores, respectively.
We provide comprehensive results for each dataset in \Cref{app: faithfulness}.}
\label{table:faithful_avg}
\end{table*}

\subsection{Desiderata of Rationales for Calibration} 
\label{sec: exp_faith}
In this section, we explore whether underlying characteristics of explanations are indicative of their usefulness to calibrators.
In \Cref{table:faithful_avg}, we report two metrics commonly used to estimate faithfulness of explanations -- \textit{sufficiency} and \textit{comprehensiveness}.
The RGF approach produces the most \textit{comprehensive} explanations across all OOD datasets when compared to CFs generated by LLMs, while in terms of \textit{sufficiency}, all CF-augmented models report higher scores compared to the RGF baseline.
As \textit{comprehensiveness} is higher when a larger number of tokens is salient for the prediction, while higher \textit{sufficiency} means that the model relies on a smaller subset of tokens, the results imply higher sufficiency of explanations is indicative of calibrator model performance. 
This is intuitive as the RGF approach generates minimally different counterfactuals with a large amount of feature overlap. On the contrary, we believe diverse CFs generated by our approach force the models to capture nuanced differences in explanations between instances. 

\section{Conclusion}
In our paper, we present a novel approach for automatic data augmentation by LLM generated counterfactual instances diverse in surface form and semantic content.
Our results show that augmenting training data of smaller models with LLM generated CFs consistently improves generalization capabilities of SLMs across six OOD extractive QA datasets.
We further show that models trained on CF augmented data are easier to calibrate, both when considering the standard confidence-based setup as well as the explanation-augmented calibration setup.
Finally, we show that rationale-augmented calibrator models prefer concise explanations, rather than comprehensive ones.
By highlighting the fact that more diverse CF instances improve the quality of the models' internal representations  
we pave the way for future works exploring the relation between surface form and semantic diversity of data used for augmentation and the models' generalization performance.

\label{sec:bibtex}

\section*{Limitations}
Our work only concentrates on the extractive QA task and can be extended to other generative tasks in the future. In addition, our approach of generating CFs can be computationally expensive for very large models and therefore we constrained ourselves to a maximum model size of 20B. In future, smaller and efficient LLMs can even make our methods better applicable. For model calibration, we utilize SHAP explanations as baselines from prior work which are also compute intensive since they need to compute many perturbations on the data. But these compute based limitations should not limit the applicability of our methods since we also show that efficient explanations based on attentions and gradients can also perform at par or sometimes even better than SHAP.

\section*{Ethics and Broader Impact Statement}
The core of our work is based on the ability of LLMs to generate reasonable explanations but prior works have shown that these models hallucinate and are not free from biases captured from large-scale web data. These hallucinations and biases might trickle down to SLM as we augment them with LLM generated CF data. To overcome these issues, we design our approaches with hard and soft filtering stages that try to eliminate such noisy and biased data and still achieve significant improvements over existing baselines.

\section*{Acknowledgements}
This work has been funded by the German Research Foundation (DFG) as part of the UKP-SQuARE project (grant GU 798/29-1), by the German Federal Ministry of Education and Research and the Hessian Ministry of Higher Education, Research, Science and the Arts within their joint support of the National Research Center for Applied Cybersecurity ATHENE, and by the European Union (ERC, InterText, 101054961). 
Views and opinions expressed are, however, those of the author(s) only and do not necessarily reflect those of the European Union or the European Research Council. Neither the European Union nor the granting authority can be held responsible for them.

\bibliography{main}

\appendix
\clearpage
\section{Datasets}
\label{app:datasets}

We perform our experiments on English language datasets.
SQuAD is a reading comprehension dataset, consisting of questions posed by crowdworkers on a set of Wikipedia articles. The SQuAD-adversarial dataset is an adversarial setting of SQuAD wherein automatically generated distractor sentences are inserted at the end of each example context to distract computer systems without changing the correct answer or misleading humans.
TriviaQA comprises of QA pairs sourced from trivia and quiz-league websites. 
Similar to SQuAD, HotpotQA is also comprised of passages extracted from wikipedia but consists of questions requiring multiple reasoning steps. 
Natural Questions consists of questions collected from information-seeking
queries to the Google search engine by real users under natural conditions.
Answers to the questions are annotated in a retrieved Wikipedia page by crowdworkers. 
NewsQA is a challenging machine comprehension dataset of human-generated question-answer pairs based on a set of over 10,000 news articles from CNN. BioASQ is a large-scale biomedical semantic indexing and question answering dataset collected by domain experts. 
For evaluation, we use the pre-processed test sets from the MRQA shared task~\cite{DBLP:conf/acl-mrqa/FischTJSCC19}.

\section{Models}
\label{app:models}
We use the instruction tuned GPT models, namely GPT-JT and GPT-NeoxT. GPT-JT is a fork of EleutherAI's GPT-J~(6B) model trained on diverse data such as chain-of-thought~\cite{DBLP:conf/nips/Wei0SBIXCLZ22}, Public Pool of Prompts~(P3) dataset~\cite{DBLP:conf/iclr/SanhWRBSACSRDBX22}, and Natural-Instructions~(NI)~\cite{DBLP:conf/emnlp/WangMAKMNADASPK22} with the UL2 training objective~\cite{tay2022ul2}. GPT-NeoxT is based on ElutherAI’s GPT-NeoX~(20B) model fine-tuned on a set of 43M high quality dialog-style interactions spanning tasks such as QA, classification, extraction, and summarization. LLaMA models are a collection of foundation language models ranging from 7B to 65B parameters trained using publicly available datasets exclusively. In our work, we specifically utilize the LLaMA~(13B) model, which aligns with the size of other language models we employ. We additionally use the Alpaca model, obtained through Low-Rank Adaptation~(LoRA)~\cite{hu2022lora} fine-tuning of the LLaMA~(13B) model on the Alpaca dataset for 10 epochs.
Lastly, we also experiment with the Fine-tuned LAnguage Net (FLAN) models. FLAN fine-tunes the model on a large set of varied instructions that use a simple and intuitive description of the task such as “Classify this movie review as positive or negative,” or “Translate this sentence to Danish.” Specifically, we use the FLAN versions of T5-11B~\cite{10.5555/3455716.3455856} and 20B UL2 (Unified Language Learner)~\cite{tay2022ul2} models.

\section{Training, Infrastructure and Runtime}

We use a server with $8$ NVIDIA A$100$ Tensor Core GPUs, each with $80$GB VRAM to run all our experiments.
Each individual experiment required at most one A$100$ GPU.
LoRA fine-tuning of the Alpaca model took $10$ hours using the refined Alpaca dataset.
Generating counterfactual instances with LLMs, in total, took $24$-$72$ hours per model and dataset.
Training each base RoBERTa model augmented with CF instances took $4$ hours, on average per dataset, while inference on OOD datasets required a few minutes per dataset.
Training the calibrator random forest model took a maximum of five minutes across all models, datasets and input feature sets.
Computing importance features of explanations for all methods except SHAP took $1$-$2$ hours per experiment, while due to the computational complexity of SHAP, each experiment required $3$-$4$ days.

We used the following hyperparameters to train the RoBERTa model used throughout our experiments: (\underline{1}) learning rate: $1$e-$5$; (\underline{2}) batch size: 64; (\underline{3}) epochs: $5$; (\underline{4}) warmup ratio: $0.06$, (\underline{5}) max input source length: $384$.
When generating counterfactual instances using LLMs, the hyperparameters used during inference were: (\underline{1}) max new tokens: $50$, (\underline{2}) temperature: $0.7$ for all considered LLMs.

\section{Prompt Details}
\label{app:prompts}
In Table~\ref{table:prompts}, we list the prompts used to generate CF data from the LLMs. 
Prompts for the \textit{GPT} and \textit{LLaMA} family are almost similar apart from minor tweaks according to the model type, e.g. \textit{GPT-NeoxT} is a chat-based model so the instruction has to follow a chat style and \textit{Alpaca} needs a specific instruction format based on its training. 
The \textit{Flan} models follow a two-stage approach of generation: the question generation prompt asks for a question that can be answered from the context in a short span of 10 words~(following SQuAD which has small answer spans) and the answer generation prompt asks for the answer to the generated question from the input context.
If the question is not answerable, we ask the model to give \textit{I don't know} as output.
Doing this maintains the fidelity of the answer generation capability of the model.

In Table~\ref{table:crf}, we list the prompt used for the context relevance filtering stage using the Flan-UL2 model. Given a generated QA pair from a context, we prompt the model to find if the context aligns with the answer and vice-versa. We ask the model to output a hard decision in terms of True or False.

\begin{table*}[htbp]
\small
\centering
\begin{tabularx}{\textwidth}{lX}
\toprule
\textbf{Model} & \textbf{Prompt} \\
\midrule

GPT-JT \& LLaMA &  As a question generator, your task is to create a concise and clear question that can be answered by an answer span within a given context. 
The context should be a piece of text, such as a news article or historical document, and the question should require understanding and analysis of the information presented in the context. 
Your generated question should focus on key details or events described in the context and should demonstrate your ability to identify important information. 
Additionally, please ensure that your question is specific enough to have a single correct answer within the given context. 
Please note that you may need to read through the provided context multiple times to fully understand its contents before generating an appropriate question. \\
&Context: {original context} \\
&Question: {original question} \\
&Answer: {original answer} \\
& \\
&Context: {context} \\
&Question: \\

\midrule

Alpaca &  Below is an instruction that describes a task, paired with an input that provides further context. 
Write a response that appropriately completes the request. \\
&  \#\#\# Instruction: \\
&As a question generator, your task is to create a concise and clear question that can be answered by an answer span within the given context. 
The context should be a piece of text, such as a news article or historical document, and the question should require understanding and analysis of the information presented in the context. 
Your generated question should focus on key details or events described in the context and should demonstrate your ability to identify important information. 
Additionally, please ensure that your question is specific enough to have a single correct answer within the given context. 
Please note that you may need to read through the provided context multiple times to fully understand its contents before generating an appropriate question.  \\
&Context: {original context} \\
&Question: {original question} \\
&Answer: {original answer} \\
& \\
&Context: {context} \\
&Question: \\

\midrule
&  Who wanted to take over North Korea? \\ 

GPT-NeoxT &  As a question generator, your task is to create a clear and concise question that can be answered by an answer span within a given context. 
The context should be a piece of text, such as a news article or historical document, and the question should require understanding and analysis of the information presented in the context. 
Your generated question should focus on key details or events described in the context, requiring readers to carefully read and analyze the provided text. 
Please ensure that your question is specific enough to have only one correct answer span within the given context. 
Please note that you should aim for clarity and concision while still maintaining accuracy and relevance to the provided text. \\
&Context: {original context} \\
&Question: {original question} \\
&Answer: {original answer} \\
& \\
&Context: {context} \\
&Question: \\

\midrule
Flan-T5 \& UL2 & \textbf{Question generation:} \\

&  Generate a fluent and answerable question from the given context. Ensure that the answer is a span in the context and is less than 10 words.\\
&Context: {original context} \\
&Question: {original question} \\
&Answer: {original answer} \\
&Context: {context} \\
&Question: \\

& \textbf{Answer generation:} \\

&  Answer the question based on the context below. If the question cannot be answered using the information provided, then answer with "I don't know".\\
&Context: {original context} \\
&Question: {original question} \\
&Answer: {original answer} \\
&Context: {context} \\
&Question: {generated question} \\
&Answer: \\

\bottomrule
\end{tabularx}
\caption{Prompts used to generate diverse counterfactual data.}
\label{table:prompts}
\end{table*}

\begin{table*}[htbp]
\small
\centering
\begin{tabularx}{\textwidth}{lX}
\toprule
\textbf{Prompt} \\
\midrule

Given the question: \\
\{generated question\} \\
Decide if the following retrieved context is relevant to the \{generated answer\}: \\
\{retrieved context\}  \\
Answer in the following format: \\
Context is relevant: True or False. \\

\bottomrule
\end{tabularx}
\caption{Prompt used for the context relevance filtering stage.}
\label{table:crf}
\end{table*}

\section{Filtering for Data Augmentation}\label{app:filtering_cfs}


\paragraph{Context Relevance Filtering.} Despite their impressive generative capabilities, LLMs are still prone to generating open-ended questions that cannot be answered from the information provided in the input context alone.
To account for such cases, in context relevance filtering, we use the FLAN-UL2 \cite{tay2022ul2} model to filter samples where the context $\hat{c}$ does not provide sufficient information for a model to correctly answer the generated question $\hat{q}$ by discarding questions which the model labeled as unanswerable (see Appendix~\ref{app:prompts}).
This approach may also discard some complex, but answerable questions, such as those based on chronological types.\footnote{We sample $100$ instances generated by LLMs and notice that such cases occur only in the Duo-QAG approach. We hypothesize that this is due to the generation methodology of the Duo-QAG approach, which, due to its looser coupling during CF generation, produces more diverse and complex questions than the Solo-QAG approach.} 
An example of one such case that consists of multiple event dates which tends to confuse the context filtering model is given below:

\noindent \textit{\textbf{Question}: \textcolor{blue!60}{\textbf{What year was the anniversary of the Cunard liner company?}}}

\noindent \textit{\textbf{Context}: In \underline{2011}, all three Cunard ships in service changed vessel registry to Hamilton, Bermuda, the first time in the 171-year history. On \colorbox{yellow}{25th May 2015}, the three Cunard ocean liners sailed up the Mersey into Liverpool to commemorate the \colorbox{yellow}{175th anniversary} of Cunard. The ships performed manoeuvres in the celebrations of the centenary of the Cunard Building on \colorbox{yellow}{2nd June 2016}.}

\paragraph{Noise Filtering.} %
To retain such answerable questions incorrectly discarded by context relevance filtering, we use the round-trip consistency approach \cite{alberti-etal-2019-synthetic, DBLP:journals/corr/abs-2009-05167} which leverages existing QA models to answer the LLM generated questions, ensuring that the predicted answer aligns with the LLM generated target answer.
During noise filtering, we employ an ensemble of three LLMs~(the same ones used to generate CFs), initialized using different random seeds during inference to verify the generated answers. The generated CFs agreed upon by at least 2 models are kept, retaining $90\%$-$95\%$ data discarded by context relevance filtering in the DuoQAG approach.

\begin{table}[!t]
\small
\centering
\begin{tabular}{llll}
\toprule
\textbf{Approach} & \textbf{Model} & \textbf{Params} & \textbf{\# CFs} \\

\midrule
RGF & T5-3B & 3B & $87$k  \\
\midrule
&GPT-JT & 6B & $73$k~($45$k)  \\
&LLaMA & 13B & $67$k~($47$k) \\
&Alpaca & 13B & $63$k~($50$k)  \\
\multirow{-4}{*}{Solo-QAG} & GPT-NeoxT &  20B & \num{76}k~(\num{44}k)  \\
\midrule
&Flan T5-xxl & 11B & \num{72}k~(\num{66}k)  \\
 \multirow{-2}{*}{Duo-QAG} & Flan-UL2 & 20B & \num{74}k~(\num{71}k)\\
\bottomrule
\end{tabular}
\caption{General information about the counterfactual generation approaches, models used, their size and the total number of generated (selected) counterfactuals.}
\label{table:cf_info}
\end{table}

\section{Generated Counterfactuals}
\label{app: gen_cf}





We report an overview of models used to generate CFs, their parameter sizes, and the resulting number of generated (usable) CFs in \Cref{table:cf_info}. The Duo-QAG approach yields a significantly higher number of usable samples~(\textasciitilde70k) compared to Solo-QAG~(\textasciitilde50k), indicating that the two-step approach produces higher fidelity CF instances.
We hypothesize that the better generative abilities of the Duo-QAG approach arise from the extensive pre-training of FLAN-based LLMs on question generation and question-answering tasks.

In \Cref{table:generations}, we highlight semantic variations for randomly sampled counterfactuals generated by our approach.
Even in the random sample, we can observe a variety of semantic changes such as metonymy, topic shift, clarification, reversal, and expansion. %

\begin{table*}[!t]
\small
\centering
\begin{tabularx}{\textwidth}{lllX}
\toprule
\textbf{Model} & \textbf{Example (\textcolor{red!50}{Q}uestion, \textcolor{blue!50}{C}ounterfactual)} & \textbf{Semantic change}  \\
\midrule
\multirow{6}{*}{RGF} & \textcolor{red!50}{Q}: To whom did the Virgin Mary allegedly appear in 1858 in Lourdes France? & \multirow{2}{*}{Paraphrase}\\ &\textcolor{blue!50}{C}: Who did the Virgin Mary appear to? & \\

& \textcolor{red!50}{Q}: What was Beyoncé's first fragrance called? & \multirow{2}{*}{Referential shift}\\ &\textcolor{blue!50}{C}: What is the name of Rihanna's first fragrance for women? & \\

& \textcolor{red!50}{Q}: How many UNESCO World Heritage Sites is Portugal home to? & \multirow{2}{*}{Referential shift}\\ &\textcolor{blue!50}{C}:  How many UNESCO World Heritage Sites does Italy have? & \\

\midrule
\multirow{6}{*}{LLaMA} & \textcolor{red!50}{Q}: What year did the legal importation of slaves end? & \multirow{2}{*}{Metonymy}\\ &\textcolor{blue!50}{C}: What year was the Act Prohibiting Importation of Slaves enacted? & \\

& \textcolor{red!50}{Q}: How many aircraft did Britain produce in 1940? & \multirow{2}{*}{Topic shift}\\ &\textcolor{blue!50}{C}: How much food was produced in Britain during the war? & \\
& \textcolor{red!50}{Q}: What was the name of the Luftwaffe plan to invade Britain? & \multirow{2}{*}{Lexical shift}\\ &\textcolor{blue!50}{C}:  What was the codename of the German invasion of Britain? & \\

\midrule
\multirow{6}{*}{GPT-NeoxT} & \textcolor{red!50}{Q}: What theater sits on Yale's campus? & \multirow{2}{*}{Referential shift} \\ &\textcolor{blue!50}{C}: Who founded the Yale Repertory Theatre?  & \\

& \textcolor{red!50}{Q}: Which islands were a part of the Spanish East Indies? & \multirow{2}{*}{Subject shift}\\ &\textcolor{blue!50}{C}: Who controlled the Spanish Empire? & \\

 & \textcolor{red!50}{Q}:  What is the time period called from which no writing can be found? & \multirow{2}{*}{Clarification}\\ &\textcolor{blue!50}{C}:  What is the time period called where early writing is not understood?\\

\midrule
\multirow{6}{*}{Flan-UL2} & \textcolor{red!50}{Q}:  Which team did Barcelona beat to win the UEFA Super Cup? & \multirow{2}{*}{Reversal} \\ &\textcolor{blue!50}{C}: Which team won the 2015 UEFA Super Cup?  & \\

 & \textcolor{red!50}{Q}: What is Lionel Messi's goal total in all competitions? & \multirow{2}{*}{Expansion} \\ &\textcolor{blue!50}{C}: How many goals did Lionel Messi score in his career?  &  \\

 & \textcolor{red!50}{Q}:  What is regarded as the greatest literary work in Old English? & \multirow{2}{*}{Reframing} \\ &\textcolor{blue!50}{C}: What is considered the heart of Old English literature?  & \\

\bottomrule
\end{tabularx}
\caption{Randomly sampled counterfactual questions listed by the model used and the corresponding semantic change introduced. The counterfactual questions from RGF are closer to the original SQuAD samples as compared to the counterfactual questions generated by our approach using generative LLMs.}
\label{table:generations}
\end{table*}

\section{Human Evaluation of Generated Counterfactuals}
\label{app:human_eval}
To measure the fluency of generated CF questions, we score the question on a scale ranging from 1-5, see \Cref{table:fluency_score}. A question with significant grammatical errors is assigned a low score whereas a well-written and comprehensible question is assigned a high score. We perform a small-scale human evaluation with one graduate candidate proficient in English and consider a question as fluent if it gets a score of three or above.

For measuring correctness, given a question, context, and answer pair, we set two criteria: (1) the question should be answerable from the given context, and (2) the answer should be a correct and a direct span from the context.  If these criteria are met, we consider the CF instance to be correct. Similar to the fluency setup, we hire a graduate candidate proficient in English to perform this evaluation.

\begin{table}[!t]
\small
\centering
\begin{tabular}{llp{4.3cm}}
\toprule
\textbf{Category} & \textbf{Score} & \textbf{Description} \\
\midrule

 Very poor & 1 & Incomprehensible with significant grammatical errors \\
 Poor & 2 &  Comprehensible but with several grammatical errors\\
 Fair & 3 & Coherent and clear with minor grammatical errors\\
 Good  & 4 & Coherent, clear and grammatically correct\\
 Excellent & 5 & Grammatically and semantically perfect\\

\bottomrule
\end{tabular}
\caption{Fine-grained scale for measuring the fluency of the generated counterfactual question. A low score indicates an incomprehensible text whereas a high score indicates a clear and coherent text without any grammatical errors.}
\label{table:fluency_score}
\end{table}

\section{Generalization of CF Augmented Models }
\label{app:cf_gen_complete}
In \Cref{table:ood_results_f1}, we report the F1 results of the RoBERTa-base model trained on the SQuAD dataset~(\textsc{Base}) and augmented with CF data on six OOD datasets. The results are comparable to the exact match scores in \Cref{table:ood_results}, and models augmented with CFs outperform all baselines across all datasets. We find that LLaMA and GPT-NeoxT, based on the Solo-QAG approach, perform best on TriviaQA and BioASQ datasets, while the Flan-UL2 model, based on the Duo-QAG approach, performs best on SQuAD-adversarial, HotpotQA, NQ, and NewsQA datasets. These results strongly advocate the importance of diverse CFs in learning robust features for enhanced OOD performance.

Overall, the FLAN-UL2 CF augmented model also has the highest average gain across all OOD datasets, with GPT-NeoxT and LLaMA closely behind. This is largely attributed to its strong performance across most of our evaluated OOD datasets.

\begin{table*}[!t]
\small
\centering
\begin{tabularx}{\textwidth}{l| *{7}{>{\centering\arraybackslash}X}| >{\centering\arraybackslash}X}
\toprule
\textbf{F1} & \textbf{SQuAD} & \textbf{SQuAD$_{Adv.}$} & \textbf{TriviaQA} & \textbf{HotpotQA} & \textbf{NQ} & \textbf{NewsQA} & \textbf{BioASQ}  & \textbf{$G_{\text{ood}}$} \\
\midrule
Base
& $91.46_{\mbox{\tiny0.05}}$ &  $72.45_{\mbox{\tiny0.95}}$ & $47.71_{\mbox{\tiny2.11}}$ & $63.79_{\mbox{\tiny0.41}}$ & $53.78_{\mbox{\tiny2.04}}$ & $57.85_{\mbox{\tiny0.89}}$ & $60.33_{\mbox{\tiny0.99}}$ & -\\
 
 \midrule
 
 \textsc{RGF}
 
 & \cellcolor{orange!20}$91.74_{\mbox{\tiny0.09}}$  &  $71.85_{\mbox{\tiny0.30}}$ & $54.41_{\mbox{\tiny0.15}}$ & $68.04_{\mbox{\tiny0.23}}$ &  $58.02_{\mbox{\tiny0.27}}$  &  \cellcolor{orange!20}$58.77_{\mbox{\tiny0.17}}$ & $61.74_{\mbox{\tiny0.18}}$ & $2.82$ \\

 \midrule
 
   \textsc{GPT-JT}   & $91.30_{\mbox{\tiny0.13}}$ & $72.83_{\mbox{\tiny0.39}}$  & \cellcolor{orange!20}$56.67_{\mbox{\tiny0.40}}$ & $66.60_{\mbox{\tiny0.37}}$  & $59.79_{\mbox{\tiny1.05}}$ & $58.14_{\mbox{\tiny0.64}}$ & \cellcolor{orange!20}$63.45_{\mbox{\tiny0.54}}$ & 3.60\\

   \textsc{LLaMA}   & $91.40_{\mbox{\tiny0.03}}$  & $73.72_{\mbox{\tiny0.75}}$ & \cellcolor{green!20}\textbf{$\mbox{57.63}_{\mbox{\tiny0.15}}$}  & $66.65_{\mbox{\tiny0.80}}$ & \cellcolor{orange!20}$61.65_{\mbox{\tiny0.68}}$ & $58.28_{\mbox{\tiny0.48}}$ & $62.65_{\mbox{\tiny1.08}}$ & $4.11$ \\
 
   \textsc{Alpaca}   & $91.73_{\mbox{\tiny0.06}}$ & $72.69_{\mbox{\tiny0.90}}$ & $51.22_{\mbox{\tiny1.67}}$  & $67.49_{\mbox{\tiny0.52}}$ & $57.00_{\mbox{\tiny2.61}}$ & $58.28_{\mbox{\tiny1.08}}$ & $61.27_{\mbox{\tiny1.04}}$ & $2.01$ \\

    \textsc{GPT-NeoxT}   & $91.31_{\mbox{\tiny0.16}}$  & \cellcolor{orange!20}$74.09_{\mbox{\tiny1.51}}$ &   $56.42_{\mbox{\tiny0.68}}$ & $68.28_{\mbox{\tiny0.82}}$ & $60.41_{\mbox{\tiny1.81}}$ & $58.10_{\mbox{\tiny0.61}}$ & \cellcolor{green!20}\textbf{$\mbox{64.07}_{\mbox{\tiny0.70}}$} & \cellcolor{orange!20}$4.24$\\
 
\midrule

  \textsc{Flan-T5-xxl} & \cellcolor{green!20}\textbf{$\mbox{91.85}_{\mbox{\tiny0.09}}$}  & $73.45_{\mbox{\tiny1.04}}$  & $53.55_{\mbox{\tiny0.91}}$ & \cellcolor{green!20}\textbf{$\mbox{69.29}_{\mbox{\tiny0.79}}$} & $61.03_{\mbox{\tiny1.01}}$ & $59.07_{\mbox{\tiny1.75}}$ & $62.08_{\mbox{\tiny0.16}}$ & $3.76$ \\

\textsc{Flan-UL2}  & $91.67_{\mbox{\tiny0.15}}$  & \cellcolor{green!20}\textbf{$\mbox{74.25}_{\mbox{\tiny1.05}}$} &  $55.17_{\mbox{\tiny1.17}}$  & \cellcolor{orange!20}$69.22_{\mbox{\tiny0.47}}$  & \cellcolor{green!20}\textbf{$\mbox{61.88}_{\mbox{\tiny1.05}}$} & \cellcolor{green!20}\textbf{$\mbox{59.17}_{\mbox{\tiny1.25}}$} & $62.73_{\mbox{\tiny0.18}}$  & \cellcolor{green!20}\textbf{$\mbox{4.42}$}\\

\bottomrule
\end{tabularx}
\caption{F1 results for RoBERTa-base model trained on the SQuAD dataset~(\textsc{Base}) and augmented with counterfactual data. All the results are averaged over 3 runs with different random seeds. The last column ($G_{\text{ood}}$) shows the average gain of models over the \textsc{Base} model on out-of-domain datasets. Numbers marked in \textcolor{green}{green}, \textbf{bold} and \textcolor{orange}{orange} colours represent the highest and second highest scores for the particular dataset and model, respectively.}
\label{table:ood_results_f1}
\end{table*}

\section{Details of Model Calibration} \label{app:calib_details}

\subsection{Heuristic Properties for Calibration}
\label{app:heuristics}

\paragraph{Properties}
In the following paragraphs, we define the heuristic properties in $V$ that are used for the calibration of QA models.

\paragraph{Input Segments} 
The input in the QA task can be decomposed into two segments: \textit{question} and \textit{context}. Each individual token is assigned a corresponding segment name to yield features: \textrm{Attributions to Question} and \textrm{Attributions to Context}.

\paragraph{POS Tags}
We leverage POS tags from the English Penn Treebank~\cite{DBLP:journals/coling/MarcusSM94} to identify token tags. \citet{ye-durrett-2022-explanations} have already shown that some tags are more important in making the predictions eg. \textit{proper nouns} in questions. Consequently, if a model fails to take into account the proper nouns in a QA pair, it may give incorrect predictions.

\paragraph{Conjunction of Groups}
We can combine the fine-grained features produced by input segments and POS tags to create high-level features that take into account the attributions of specific tags in the question or context of the QA pair. An example of this feature can be \textrm{Attributions of NNP in Question}.

\subsection{Including Dense Features from Explanations}
As stated in \Cref{sec:our_calib}, when considering the most important tokens based on importance scores, we use a higher percentage of answer tokens. 
This is due to our initial experiments, where such a choice had a high correlation with calibration performance.
Consequently, we exclude explanation-based features from the \textit{question} as we observed diminishing calibrator performance on their inclusion. 
We hypothesize that this decrease is due to the noise or irrelevant information introduced by the \textit{question} features. %

\subsection{Hyperparameters}
\label{app:hyperparams}
We use the RandomForest implementation from Scikit-Learn~\cite{DBLP:journals/jmlr/PedregosaVGMTGBPWDVPCBPD11}. Following the work of \citet{ye-durrett-2022-explanations}, we choose a value of 300 for the \textit{n\_estimators} parameter and a value of 20 for the \textit{max\_depth} parameter for our experiments with different explanation methods. For further experiments with the addition of dense rationale features, we determine the hyperparameters through grid search using 500 training samples and 100 validation samples. The choices of \textit{n\_estimators} are [300, 400, 500] and \textit{max\_depth} is set to 20. Based on the results, we use \textit{n\_estimators} as 500 and \textit{max\_depth} as 20 for training the rationale-augmented classifier.

\subsection{Choice of calibration metric}
\label{app:cal_metric}

Calibration metrics help assess the alignment between a model's predicted probabilities and it's observed predictions. When evaluating model calibration, the \textbf{Expected Calibration Error (ECE)} \citep{10.5555/3305381.3305518} has been widely used in prior works \citep{park-caragea-2022-calibration, DBLP:conf/emnlp/LiHC22}. The ECE can be computed by partitioning model predictions into $K$ equal sized bins according to their model confidence scores. Mathematically, the calibration error can be written as

\begin{equation*}
    {\hat{\mathcal{E}}}_k = \frac{1}{\lvert B_k \rvert} {\Bigl\lvert \sum_{i \in B_k} [\vmathbb{1} (\hat{y_i} = y_i) - \text{Conf}(x_i, \hat{y_i})] \Bigl\rvert} ,
\end{equation*}

where $x$ is the input, $y$ the ground truth,
$\hat{y}$ the prediction, and $Conf(x, \hat{y})$ is the model confidence for i-th example, and $B_k$ denotes the bin with prediction confidences bounded
between $l_k$ and $u_k$. The ECE can now be computed as the weighted average of all the bins such as

\begin{equation*}
    \text{ECE} = \sum_{k=1}^K \frac{\lvert B_k \rvert}{n} {\hat{\mathcal{E}}}_k ,
\end{equation*}

where $n$ is the number of model predictions. The goal is to minimize the ECE without diminishing accuracy. Though widely used, ECE has certain shortcomings. First, most instances are assigned similar confidence, which does not give a proper indication of correct or wrong predictions whereas an ideal calibration metric should be able to do so. Second, bucketing causes cancellation effects, ignoring instance-level calibration error as many predictions are clustered in the same buckets. As a result, there are many over-confident and under-confident predictions in the same bucket and are averaged to become closer to the average accuracy. Due to these issues, we use an enhancement of ECE called \textbf{Macro-average Calibration Error, MacroCE} \citep{si-etal-2022-examining} which considers instance-level errors, but it takes equal consideration of correct and wrong predictions made by the model. Specifically, it calculates the macro-average over calibration errors for positive and negative predictions:

\begin{align*}
\begin{split}
    \text{ICE}_{pos} ={}& \frac{1}{n_p} \sum_{i=1}^{n_p} (1- \text{Conf}(x_i, \Tilde{y_i})), \forall \Tilde{y_i} = y_i,
\end{split}\\
\begin{split}\label{eq:2}
    \text{ICE}_{neg} ={}& \frac{1}{n_n} \sum_{i=1}^{n_n} (\text{Conf}(x_i, \Tilde{y_i}) - 0), \forall \Tilde{y_i} \neq y_i, 
\end{split}\\
    \text{MacroCE} ={}& \frac{1}{2} \phantom{0} \cdot (ICE_{pos} + ICE_{neg}),
\end{align*}

where $n_p$ and $n_n$ are the correct and wrong predictions.


        

\begin{table*}[htbp]
\small
\centering
\begin{tabularx}{\textwidth}{@{}lllllllllllX@{}}
\toprule
&\multirow{2}[3]{*}{Approach} & \multicolumn{3}{c}{\textbf{SQuAD Adv.}} & \multicolumn{3}{c}{\textbf{Trivia QA}} & \multicolumn{3}{c}{\textbf{Hotpot QA}} \\
\cmidrule(lr){3-5} \cmidrule(lr){6-8} \cmidrule(lr){9-11} 
 && ACC($\uparrow$) & AUC($\uparrow$) & MCE($\downarrow$) & ACC($\uparrow$) & AUC($\uparrow$)  & MCE($\downarrow$) & ACC($\uparrow$) & AUC($\uparrow$)  & MCE($\downarrow$) \\

\midrule

&Base &   64.2 & 68.5 & 0.474 & 60.1 & 58.3 & 0.539 & 61.1 & 74.0 &  0.502 \\

&RGF &  62.2 & 67.3  & 0.480  & 65.3  & 70.5  &  0.532 & 63.0  &  79.1 & 0.493 \\

&LLaMA & 65.6  & 71.6  & 0.467  & 66.9  &  \textbf{74.8} &  \textbf{0.507} & 64.3  & 80.0  & 0.487 \\

 \rot{\rlap{~CONF}}
&GPT-NeoxT &  66.0 & 71.3 & 0.473  &  \textbf{67.8} &  74.5 &  \textbf{0.507} & 63.0  & 80.0  & \textbf{0.482} \\

&FLAN-UL2 & \textbf{66.6}  & \textbf{74.9} & \textbf{0.454}  & 66.5  & 73.2  & 0.520  & \textbf{65.1}  &  \textbf{81.6} & 0.488 \\

\midrule

&Base & 75.0 & 84.3 & 0.471  &  \textbf{72.0} &  71.8 & 0.545  & 63.3  & 75.5  & 0.504 \\

&RGF &  72.7 &  83.0 &  0.474  &  70.7  & 74.2  &  0.525 &  64.0 & 79.7  & 0.491 \\

&LLaMA &  73.4 & 84.0  &  0.468  &  70.7 & 76.1  &  \textbf{0.505} & 65.3  &  80.0 & 0.493 \\
&LLaMA + $F_{r}$ & 77.5  & 87.1 & 0.461  &  70.7 &  \textbf{76.4} & 0.506  &  \textbf{65.5} &  80.1 & 0.494 \\

&GPT-NeoxT &  74.5 & 84.7 &  0.470 &  70.3 & 75.5  & 0.508  & 62.7  & 79.6  & 0.493 \\
 \rot{\rlap{~SHAP}}
&GPT-NeoxT + $F_{r}$ & 78.2  & 87.5 &  0.461 &  70.2 & 75.6  & 0.509  &  63.0 & 79.7  & 0.493\\

&FLAN-UL2 & 75.2  & 85.4 & 0.468  &  70.7 & 75.3  & 0.515  & 65.3  &  81.1 & \textbf{0.489} \\
&FLAN-UL2 + $F_{r}$ & \textbf{79.1}  & \textbf{88.4} &  \textbf{0.460} &  70.8 &  75.3 & 0.517  &  65.4 & \textbf{81.2}  & 0.490 \\

\midrule

&Base & 65.7  & 78.1 & 0.476  &  \textbf{73.5} &  71.9 & 0.558  &  63.1 &  74.5 & 0.509 \\

&RGF &  64.1 & 77.6 &  0.474 &  70.3 & 73.9  & 0.533  & 63.3  & 78.9  & 0.495 \\

&LLaMA &  65.4  & 78.1 & 0.475  & 70.5  & 76.3  &  \textbf{0.510} &  64.0 &  78.9 & 0.493\\
&LLaMA + $F_{r}$ &  71.6 & 84.5 &  0.461 &  70.5 & \textbf{76.6}  &  0.511 & 64.2  & 79.1  & 0.494 \\

&GPT-NeoxT &  67.0 & 79.0 &  0.475 & 69.9  & 75.6  & 0.511  & 62.5  & 78.7  & 0.492 \\
&GPT-NeoxT + $F_{r}$ &  72.9  & 84.9 & \textbf{0.459}  & 70.1  &  75.8 &  0.511 &  62.6 & 78.8  & 0.494 \\

 \rot{\rlap{~SC. ATTN.}}
&FLAN-UL2 &  67.2 & 79.1 & 0.479  &  70.7 &  75.2 & 0.520  & 64.2  &  80.0  & \textbf{0.491} \\
&FLAN-UL2 + $F_{r}$ &  \textbf{73.2} & \textbf{85.8} & 0.461  &  70.8 & 75.6  & 0.521  & \textbf{64.3}  & \textbf{80.3}  & \textbf{0.491} \\

\midrule

&Base & 65.5  & 77.9 & 0.476  &  \textbf{72.9} &  72.2 & 0.554  & 62.9  & 74.6  & 0.507 \\

&RGF & 64.2  & 76.9 &  0.477 &  70.7 & 74.3  & 0.530  &  63.3 & 78.7  & 0.495 \\

&LLaMA &  66.5  & 79.1 & 0.474  &  70.5 & 76.0  & \textbf{0.508}  & 64.1  &  78.9 &  0.493 \\
&LLaMA + $F_{r}$ &  72.4 & 84.5 & 0.464  & 70.6  &  \textbf{76.3} &  0.510 &  \textbf{64.4} & 79.1  & 0.494 \\

 \rot{\rlap{~IG}}
 
&GPT-NeoxT & 66.6  & 79.2 & 0.474  &  70.2 & 75.2  & 0.511  & 62.0  & 78.5  & 0.493 \\
&GPT-NeoxT + $F_{r}$ & 73.0  & 84.7 &  \textbf{0.462} & 70.2  & 75.6  & 0.511  & 61.9  & 78.3  & 0.495\\

&FLAN-UL2 & 68.7 & 81.0 & 0.475  & 70.6  &  74.9 &  0.521 & 64.2  &  79.9 & \textbf{0.492} \\
&FLAN-UL2 + $F_{r}$ & \textbf{74.6}  & \textbf{85.8} & 0.464  & 70.8  & 75.3  & 0.524  &  64.3 & \textbf{80.4}  & \textbf{0.492}\\

\midrule
\midrule

&Base &   66.0 & 77.8 & 0.477 & \textbf{71.7} & 70.9 & 0.557 & 62.4 & 73.9 &  0.508 \\

&RGF & 63.7 & 76.0 & 0.479 & 70.5 & 73.8 & 0.532 & 63.2 & 78.7 & 0.494 \\

&LLaMA & 64.7 & 77.4 &  0.477 & 70.4 & 76.1 & \textbf{0.508} & 64.0 & 79.1 & 0.493  \\
&LLaMA + $F_{r}$ & 71.4 & 84.0 &  \textbf{0.466} & 70.4 & \textbf{76.3} & 0.510 & \textbf{64.3} & 79.4 & 0.493 \\

&GPT-NeoxT & 66.2 & 78.3 &  0.477 & 70.0 & 74.8 & 0.511 & 61.9 & 78.4 & 0.493 \\
 \rot{\rlap{~ATTN.}}
&GPT-NeoxT + $F_{r}$ & 72.2  & 84.4 &  \textbf{0.466} & 69.8 & 74.9 & 0.511 & 62.0 & 78.4 & 0.494 \\

&FLAN-UL2 & 66.3 & 78.6 &  0.478 & 70.4 & 74.9 & 0.519 & 64.0 & 79.9 & \textbf{0.492} \\
&FLAN-UL2 + $F_{r}$ & \textbf{72.5} & \textbf{84.9} &  0.467 & 70.5 & 75.2 & 0.521 & 64.1 & \textbf{80.2} & \textbf{0.492} \\

\midrule

&Base &  65.1 & 78.1  & 0.474 & \textbf{72.3} & 71.4  & 0.556 & 63.0 & 74.2 & 0.508 \\

 &RGF  & 63.2 & 76.3 & 0.478 & 70.5 & 74.0  & 0.531 & 63.2 & 78.7 & 0.495 \\

 &LLaMA & 65.7 & 78.5 & 0.475 & 70.5 & 76.2 & \textbf{0.509} & 64.2 & 78.8 & 0.494 \\
 &LLAMA + $F_{r}$ & 72.1 & 84.6 &  0.464 & 70.6 & \textbf{76.3} & 0.510 & 64.2 & 79.0 & 0.494 \\

& GPT-NeoxT & 66.3 & 78.9 &  0.475 & 70.1 & 75.2 & 0.511 & 62.1 & 78.5 & 0.493 \\
& GPT-NeoxT + $F_{r}$ & 72.7 & 84.9 &  \textbf{0.461} & 70.3 & 75.5 & 0.511 & 62.3 & 78.5 & 0.495 \\
 
 \rot{\rlap{~INP$X$GRAD}}

&FLAN-UL2 & 67.0 & 79.5 &  0.477 & 70.7 & 74.9 & 0.520 & 64.2 & 80.1 & \textbf{0.492} \\

&FLAN-UL2 + $F_{r}$ & \textbf{73.0} & \textbf{85.7} &  0.462 & 70.7 & 75.1 & 0.522 & \textbf{64.3} & \textbf{80.3} & 0.493 \\

\bottomrule
\end{tabularx}
\caption{Calibration results for a Roberta-base model trained on SQuAD when transferring to out-of-domain settings using explanations based on attention and gradients.}
\label{table:calib_remaining_1}
\end{table*}

\begin{table*}[htbp]
\small
\centering
\begin{tabularx}{\textwidth}{@{}lllllllllllX@{}}
\toprule
&\multirow{2}[3]{*}{Approach} & \multicolumn{3}{c}{\textbf{NQ}} & \multicolumn{3}{c}{\textbf{News QA}} & \multicolumn{3}{c}{\textbf{BioASQ}} \\
\cmidrule(lr){3-5} \cmidrule(lr){6-8} \cmidrule(lr){9-11} 
 && ACC($\uparrow$) & AUC($\uparrow$) & MCE($\downarrow$) & ACC($\uparrow$) & AUC($\uparrow$)  & MCE($\downarrow$) & ACC($\uparrow$) & AUC($\uparrow$)  & MCE($\downarrow$) \\

\midrule

&Base &   68.6 & 70.1 & 0.531 & 68.0 & 71.3 & 0.535 & 63.3 & 72.1 &  0.499 \\

&RGF&   69.8 & 75.4 & 0.510 & 68.5 & 75.8 & 0.538  & 66.3 & 76.3 &  0.498 \\

&LLaMA &   \textbf{72.7} & 80.5 & 0.490 & 69.9 & 75.1 & 0.530 & 66.8 & 79.5 &  0.477 \\

 \rot{\rlap{~CONF}}
&GPT-NeoxT &   72.0 & \textbf{81.3} & \textbf{0.476} & \textbf{70.3} & 75.1 & 0.534 & 65.1 & \textbf{80.2} &  \textbf{0.465} \\

&FLAN-UL2 &   70.6 & 80.0 & 0.494 & 68.7 & \textbf{76.7} & \textbf{0.527} & \textbf{67.5} & 77.0 &  0.494 \\

\midrule

&Base &   \textbf{76.9} & 77.9 & 0.520 & 70.9 & 75.2 & 0.517 & 69.9 & 76.8 &  0.504 \\

&RGF &   75.3 & 79.8 & 0.507 & 69.5 & 75.2 & 0.512 & 69.9 & 78.8 &  0.495 \\

&LLaMA &   73.8 & 80.3 & 0.484 & 70.0 & 75.3 & 0.517 & 72.1 & 82.2 &  0.493 \\
&LLaMA + $F_{r}$ &   73.8 & 80.5 & 0.486 & 70.0 & 75.3 & 0.516 & 72.0 & 82.2 &  0.493 \\

&GPT-NeoxT &   73.0 & 81.0 & \textbf{0.478} & 69.8 & 74.0 & 0.518 & 71.0 & \textbf{82.4} &  \textbf{0.485} \\
 \rot{\rlap{~SHAP}}
&GPT-NeoxT + $F_{r}$ &   73.1 & 81.2 & 0.480 & 69.8 & 74.0 & 0.517 & 69.7 & 82.3 &  \textbf{0.485} \\

&FLAN-UL2 &   73.9 & 81.4 & 0.493 & 71.1 & 77.4 & \textbf{0.514} & \textbf{73.2} & 81.0 &  0.497 \\
&FLAN-UL2 + $F_{r}$ &   74.1 & \textbf{81.5} & 0.494 & \textbf{71.2} & \textbf{77.5} & \textbf{0.514} & 72.1 & 81.0 &  0.497 \\

\midrule

&Base &   \textbf{76.6} & 77.7 & 0.526 & \textbf{72.0} & 76.2 & 0.538 & 71.7 & 78.1 &  0.509 \\

&RGF &   75.2 & 79.7 & 0.511 & 69.8 & 75.8 & 0.523 & 70.8 & 79.0 &  0.503 \\

&LLaMA &   74.0 & 80.4 & 0.489 & 70.4 & 75.6 & 0.528 & 72.7 & 82.5 &  0.494 \\
&LLaMA + $F_{r}$ &   74.1 & 80.5 & 0.490 & 70.5 & 75.8 & 0.529 & 72.4 & 82.3 &  0.495 \\

&GPT-NeoxT &   72.7 & 80.9 & \textbf{0.481} & 70.3 & 74.5 & \textbf{0.526} & 72.9 & 83.4 &  \textbf{0.486} \\
&GPT-NeoxT + $F_{r}$ &   72.9 & 81.0 & 0.482 & 70.2 & 74.8 & \textbf{0.526} & 72.0 & 83.0 &  0.487 \\

 \rot{\rlap{~SC. ATTN.}}
&FLAN-UL2 &   74.5 & 81.3 & 0.495 & 71.3 & 77.7 & 0.527 & \textbf{73.7} & 81.0 &  0.502 \\
&FLAN-UL2 + $F_{r}$ &   74.4 & \textbf{81.4} & 0.496 & 71.6 & \textbf{78.1} & 0.528 & 73.3 & \textbf{81.3} &  0.501 \\

\midrule

&Base &   \textbf{76.7} & 77.7 & 0.526 & \textbf{72.0} & 76.2 & 0.536 & 70.4 & 77.0 &  0.507 \\

&RGF &   75.3 & 79.7 & 0.511 & 69.9 & 75.9 & 0.519 & 70.3 & 78.7 &  0.498 \\

&LLaMA &   74.0 & 80.4 & 0.489 & 70.6 & 75.8 & 0.526 & 71.6 & 81.9 &  0.495 \\
&LLaMA + $F_{r}$ &   74.0 & 80.5 & 0.490 & 70.5 & 76.0 & 0.527 & 72.0 & 82.2 &  0.495 \\

 \rot{\rlap{~IG}}
 
&GPT-NeoxT &   73.0 & 80.7 & \textbf{0.482} & 70.0 & 74.3 & 0.525 & 71.6 & 82.4 &  0.488 \\
&GPT-NeoxT + $F_{r}$ &   73.0 & 80.8 & 0.483 & 70.2 & 74.6 & \textbf{0.524} & 70.3 & \textbf{82.5} &  \textbf{0.487} \\

&FLAN-UL2 &   74.2 & 81.3 & 0.496 & 71.4 & 77.9 & 0.528 & \textbf{73.1} & 80.7 &  0.502 \\
&FLAN-UL2 + $F_{r}$ &   74.3 & \textbf{81.4} & 0.497 & 71.7 & \textbf{78.3} & 0.529 & 73.0 & 80.9 &  0.501 \\

\midrule
\midrule

&Base &   \textbf{76.6} & 77.5 & 0.527 & \textbf{71.5} & 75.4 & 0.536 & 70.9 & 77.3 &  0.509 \\

&RGF &   75.4 & 79.8 & 0.511 & 70.1 & 75.6 & 0.520 & 70.6 & 78.6 &  0.502 \\

&LLaMA &   73.9 & 80.4 & 0.490 & 70.2 & 75.5 & 0.526 & 72.6 & 82.6 &  0.494 \\
&LLaMA + $F_{r}$ &   74.0 & 80.6 & 0.490 & 70.4 & 75.7 & 0.527 & 72.2 & 82.4 &  0.495 \\

&GPT-NeoxT &   72.9 & 80.8 & 0.481 & 69.7 & 74.1 & 0.524 & 71.7 & \textbf{82.7} &  \textbf{0.485} \\
 \rot{\rlap{~ATTN.}}
&GPT-NeoxT + $F_{r}$ &   73.1 & 81.0 & 0.482 & 69.8 & 74.4 & \textbf{0.523} & 70.5 & 82.6 &  0.486 \\

&FLAN-UL2 &   74.1 & 81.2 & \textbf{0.496} & 71.2 & 77.5 & 0.526 & \textbf{73.0} & 80.7 &  0.503 \\
&FLAN-UL2 + $F_{r}$ &   74.2 & \textbf{81.4} & 0.497 & 71.4 & \textbf{77.8} & 0.528 & 71.9 & 80.2 &  0.504 \\

\midrule

&Base &   \textbf{76.9} & 77.9 & 0.524 & \textbf{71.9} & 76.0 & 0.534 & 70.7 & 76.8 &  0.508 \\
 &RGF  &   75.7 & 79.9 & 0.510 & 70.1 & 75.7 & 0.520 & 70.4 & 78.5 &  0.500 \\
 &LLaMA &   73.8 & 80.4 & 0.488 & 70.3 & 75.7 & 0.525 & 71.8 & 82.0 &  0.495 \\
 &LLAMA + $F_{r}$ &   73.8 & 80.5 & 0.490 & 70.5 & 76.0 & 0.526 & 71.7 & 81.9 &  0.495 \\

& GPT-NeoxT &   72.8 & 80.9 & \textbf{0.481} & 70.2 & 74.4 & \textbf{0.523} & 71.7 & 82.8 &  \textbf{0.487} \\
& GPT-NeoxT + $F_{r}$ &   72.9 & 81.0 & 0.482 & 70.2 & 74.6 & 0.524 & 70.9 & \textbf{83.0} &  \textbf{0.487} \\
 
 \rot{\rlap{~INP$X$GRAD}}

&FLAN-UL2 &   74.2 & 81.4 & 0.494 & 71.2 & 77.7 & 0.527 & 73.0 & 80.7 &  0.503 \\
&FLAN-UL2 + $F_{r}$ &   74.2 & \textbf{81.5} & 0.496 & 71.6 & \textbf{78.1} & 0.528 &\textbf{73.1} & 80.8 &  0.502 \\

\bottomrule
\end{tabularx}
\caption{\textit{Cont.} Calibration results for a Roberta-base model trained on SQuAD when transferring to out-of-domain settings using explanations based on attention and gradients.}
\label{table:calib_remaining_2}
\end{table*}

\section{Additional Results}
\subsection{Model Calibration}
\label{app:model_calib}

\begin{figure*}[!t]
\begin{subfigure}{\textwidth}
    \includegraphics[width=\linewidth]{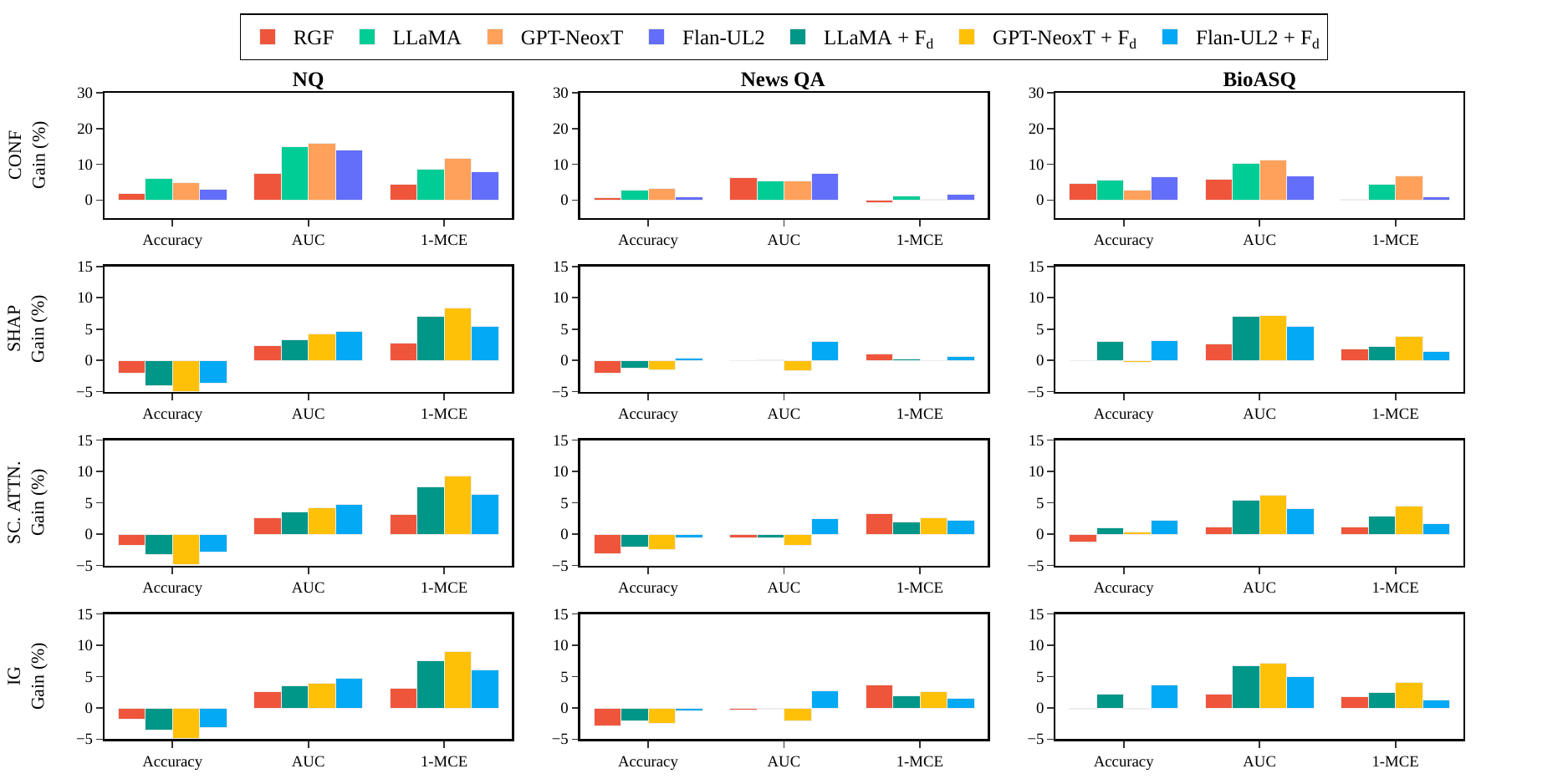}
\end{subfigure}\hfill

\caption{Percentage improvement of CF-augmented models' calibration performance over the unaugmented RoBERTa-base model trained on SQuAD, using features based on probability~(\textsc{conf}) and rationales from \textsc{shap}, scaled attention and integrated gradients.
The results for \textsc{conf} (row \#1) are reported on models which do not use explanation-based features.
In the remaining experiments (other rows), along with \textsc{base} and \textsc{rgf}, we report the results of dense-feature augmented calibrators.} 
\label{fig:calib_results_2}
\end{figure*}

We report the model calibration results as \% improvement over the base model in \Cref{fig:calib_results_2} for the NQ, NewsQA, and BioASQ datasets.
We compare our models against two baselines: (1) \textsc{conf}, where the calibrator model only uses the thresholded probability of the predicted class to assess whether the prediction is trustworthy, and (2) \textsc{shap}. %
On the \textsc{conf} baseline, when only the probability of the underlying model is used as input to the calibrator, our CF-augmented models improve calibration accuracy across all OOD datasets with an average increase of \textasciitilde$5\%$.
These results suggest that augmenting a model with counterfactual instances already improves the model's capability to capture nuanced shifts in the data distribution. 
Improved robustness of CF-augmented models is further evident from the high inverse MacroCE scores on the \textsc{conf} baseline where even without features from explanations, CF-augmented models exhibit the best calibration scores~(\textasciitilde$+6\%$) across all datasets. 

When incorporating explanation features, on the \textsc{shap} baseline, the CF-augmented models improve calibration accuracy by an average of \textasciitilde$3\%$ on two out of three OOD datasets, the exception being NQ, where the accuracy decreases marginally. 
Nevertheless, the CF-augmented models achieve superior AUC scores on all OOD datasets with an average improvement of \textasciitilde$5.5\%$ compared to the SHAP baseline without CF augmentation. 
For completeness, we report results produced by $\alpha$ and $x\nabla x$ in the \Cref{app:model_calib}. 

Overall, the CF-augmented models coupled with dense rationale features improve calibration over all baselines, all explanation methods, and OOD datasets.
Our results show that augmenting training data with CF instances improves model calibration and that calibrators benefit from the semantic content of the most salient tokens from explanations. %

In \Cref{table:calib_remaining_1} and \Cref{table:calib_remaining_2}, we present the results of the calibration of LLM augmented models using features based on probability (CONF), and heuristics from three explanation methods: SHAP, scaled attention, and integrated gradients. Additionally, we include explanations based on \textit{attention} and \textit{input$X$gradients}. For these additional explanation methods, we observe that the Flan-UL2 performs best on SQuAD adversarial, HotpotQA, and BioASQ datasets. On NQ and NewsQA datasets, our augmented models do not outperform the base model on accuracy but show significant improvements on AUC with a gain of ~\textasciitilde+4 and ~\textasciitilde+2 points, respectively using the Flan-UL2 CFs. Similarly, for TriviaQA, the LLaMA CFs improve the AUC by ~\textasciitilde+5 points over the base model. These results indicate that our calibration methodology helps in calibrating models across a wide range of evaluated explanations.

\subsection{Desiderata of Explanations for Calibration}
\label{app: faithfulness}
In \Cref{table:faithful}, we report the comprehensiveness and sufficiency scores of explanations generated by the baseline and CF-augmented models for all the OOD datasets under evaluation. The RGF baseline provides the most comprehensive explanations across all datasets and explanation methods when compared to our CF-augmented models, while in terms of sufficiency, we observe an opposite trend with the RGF baseline performing worse than all the CF-augmented models. We hypothesize that the CF augmented models produce highly sufficient explanations i.e. assign more importance to a smaller subset of tokens since the diverse CFs help the base model in discerning specific key input features important for the prediction -- resulting in better OOD and calibration performance.

\begin{table*}[!t]
\small
\centering
\setlength{\tabcolsep}{10pt}
\begin{tabularx}{\textwidth}{@{}lllllll|llllX}
\toprule
\multirow{2}[3]{*}{} & \multirow{2}[3]{*}{Model} & \multicolumn{5}{c}{\textbf{Comprehensiveness~($\uparrow$)}} & \multicolumn{5}{c}{\textbf{Sufficiency~($\uparrow$)}} \\
\cmidrule(lr){3-7} \cmidrule(lr){8-12}  
 && $\alpha$ & $\alpha\nabla\alpha$ & x$\nabla$x & IG & SHAP & $\alpha$ & $\alpha\nabla\alpha$ & x$\nabla$x & IG & SHAP\\
\midrule

& Base & 0.33 & 0.35 &  0.37 & 0.38 & 0.34 & 0.39 & 0.39 & 0.39 & 0.39 & 0.40 \\
& RGF & \textbf{0.34} & \textbf{0.39} &  \textbf{0.38} & \textbf{0.40} & \textbf{0.37} & 0.36 & 0.38 & 0.37 & 0.37 & 0.36 \\
 & LLaMA & \textbf{0.34} & 0.36 & 0.36 & 0.36 & 0.33 & 0.40 & 0.41 & 0.40 & 0.40 & 0.41 \\
 & GPT-Neox & 0.30 & 0.35 &  0.36 & 0.36 & 0.32 & \textbf{0.41} & \textbf{0.42} & \textbf{0.42} & \textbf{0.42} & \textbf{0.43} \\
\rot{\rlap{~Squad Adv.}}
 & Flan-UL2 & 0.32 & 0.36 &  0.37 & 0.36 & 0.33 & 0.39 & 0.39 & 0.39 & 0.39 & 0.40 \\
\midrule
& Base & 0.35 & 0.35 &  0.37 & 0.39 & 0.35 & 0.56 & 0.55 & 0.56 & 0.56 & 0.54 \\
& RGF & \textbf{0.37} & \textbf{0.43} &  \textbf{0.45} & \textbf{0.47} & \textbf{0.44} & 0.43 & 0.44 & 0.44 & 0.45 & 0.43 \\
& LLaMA & 0.29 & 0.33 & 0.33 & 0.34 & 0.29 & 0.57 & 0.59 & 0.58 & 0.59 & 0.58 \\
& GPT-Neox & 0.26 & 0.29 & 0.29 & 0.30 & 0.25 & \textbf{0.63} & \textbf{0.64} & \textbf{0.64} & \textbf{0.64} & \textbf{0.63} \\
\rot{\rlap{~Trivia QA}}
& Flan-UL2 & 0.32 & 0.36 & 0.37 & 0.39 & 0.33 & 0.51 & 0.51 & 0.52 & 0.53 & 0.51 \\
\midrule
& Base & 0.29 & 0.29 & 0.32 & 0.34 & 0.31 & 0.52 & 0.52 & 0.53 & 0.53 & 0.52 \\
& RGF & 0.33 & \textbf{0.38} & \textbf{0.39} & \textbf{0.41} & \textbf{0.57} & 0.37 & 0.38 & 0.37 & 0.37 & 0.37 \\
& LLaMA & 0.30 & 0.31 & 0.31 & 0.31  & 0.29 & 0.51 & 0.52 & 0.51 & 0.51 & 0.53 \\
& GPT-Neox & 0.28 & 0.29 & 0.28 & 0.30 & 0.27 & \textbf{0.56} & \textbf{0.57} &\textbf{0.57} & \textbf{0.57} & \textbf{0.58} \\
\rot{\rlap{~Hotpot QA}}
& Flan-UL2 & \textbf{0.35} & 0.34 & 0.35 & 0.36 & 0.33 & 0.43 & 0.43 & 0.43 & 0.43 & 0.44 \\

\midrule
& Base & 0.35 & 0.34 & 0.36 & 0.37 & 0.34 & 0.55 & 0.55 & 0.55 & 0.55 & 0.55 \\
& RGF  & 0.37 & \textbf{0.39} & \textbf{0.43} & \textbf{0.43} & \textbf{0.38} & 0.46 & 0.46 & 0.47 & 0.47 & 0.46 \\
& LLaMA & 0.36 & 0.35 & 0.34 & 0.35 & 0.32 & \textbf{0.56} & \textbf{0.56} & 0.55 & 0.55 & \textbf{0.55} \\
\rot{\rlap{~NQ}}
& GPT-Neox & 0.35 & 0.32 & 0.33 & 0.34 & 0.31 & \textbf{0.56} & \textbf{0.56} & \textbf{0.56} &\textbf{0.56} & \textbf{0.55} \\
& Flan-UL2 & \textbf{0.40} & 0.37 & 0.38 & 0.38 & 0.35 & 0.49 & 0.50 & 0.49 & 0.49 & 0.49 \\

\midrule
& Base & 0.34 & 0.35 & 0.37 & 0.40 & 0.43 & 0.55 & 0.55 & 0.56 & 0.57 & 0.54\\
& RGF & \textbf{0.36} &\textbf{0.41} & \textbf{0.41} & \textbf{0.43} & \textbf{0.49} & 0.45 & 0.48 & 0.46 & 0.46 & 0.45 \\
& LLaMA & 0.30 & 0.32 & 0.32 & 0.33 & 0.37 & 0.61 & 0.62 & 0.61 & 0.62 & 0.60 \\
& GPT-Neox & 0.28 & 0.29 & 0.30 & 0.31 & 0.35 & \textbf{0.63} & \textbf{0.64} & \textbf{0.64} & \textbf{0.64} & \textbf{0.62} \\
\rot{\rlap{~News QA}}
& Flan-UL2 & 0.32 & 0.37 & 0.37 & 0.40 & 0.45 & 0.52 & 0.54 & 0.53 & 0.54 & 0.52 \\

\midrule
& Base & \textbf{0.32} & 0.38 & 0.38 & 0.40 & 0.35 & 0.50 & 0.51 & 0.51 & 0.51 & 0.50 \\
& RGF & 0.31 & \textbf{0.44} & \textbf{0.41} & \textbf{0.42} & \textbf{0.38} & 0.41 & 0.45 & 0.43 & 0.43 & 0.41 \\
& LLaMA & 0.30 & 0.34 & 0.33 & 0.34 & 0.32 & \textbf{0.57} & \textbf{0.58} & \textbf{0.57} & \textbf{0.58} & \textbf{0.57} \\
& GPT-Neox & 0.24 & 0.33 & 0.32 & 0.34 & 0.30 & 0.56 & \textbf{0.58} & \textbf{0.57} & \textbf{0.58} & \textbf{0.57} \\
\rot{\rlap{~BioASQ}}
& Flan-UL2 & 0.29 & 0.38 & 0.38 & 0.39 & 0.35 & 0.49 & 0.50 & 0.50 & 0.50 & 0.49 \\

\bottomrule
\end{tabularx}
\caption{Comprehensiveness and sufficiency scores of explanations generated by baseline and counterfactual augmented models. 
Numbers marked in \textbf{bold} represent the highest scores for the particular dataset with a corresponding model and explanation.
}
\label{table:faithful}
\end{table*}

\end{document}